\documentclass{article}

\usepackage{iclr2025_conference,times}

\usepackage[utf8]{inputenc} 
\usepackage[T1]{fontenc}    
\usepackage{hyperref}       
\usepackage{url}            
\usepackage{booktabs}       
\usepackage{amsfonts}       
\usepackage{nicefrac}       
\usepackage{microtype}      
\usepackage{xcolor}         
\usepackage{natbib}
\usepackage{longtable}
\usepackage{amsmath}
\usepackage{amssymb}
\usepackage{caption}
\usepackage{graphicx}
\usepackage{cleveref}
\usepackage{mathtools}
\usepackage{textcomp}
\usepackage{todonotes}
\usepackage{amsthm}
\usepackage{float}
\usepackage{import}
\usepackage{tikz}
\usepackage{pgfplots}
\usepackage{pgfplotstable}
\pgfplotsset{compat=1.18}
\usepackage{threeparttable}

\title{Physiome-ODE: A Benchmark for Irregularly Sampled Multivariate Time Series Forecasting
Based on Biological ODEs}
\author{%
  Christian Klötergens \\
  ISMLL \& VWFS DARC\\
  University of Hildesheim\\
  Hildesheim, Germany\\
  \texttt{kloetergens@ismll.de}\\
  \And
  Vijaya Krishna Yalavarthi\\
  ISMLL\\
  University of Hildesheim\\
  Hildesheim, Germany\\
  \And
  Randolf Scholz\\
  ISMLL\\
  University of Hildesheim\\
  Hildesheim, Germany\\
  \And
  Maximilian Stubbemann\\
  ISMLL \& VWFS DARC\\
  University of Hildesheim\\
  Hildesheim, Germany\\
  \And
  Stefan Born\\
  Institute of Mathematics\\
  TU Berlin\\
  Berlin, Germany\\
  \And
  Lars Schmidt-Thieme\\
  ISMLL \& VWFS DARC\\
  University of Hildesheim\\
  Hildesheim, Germany\\
}

\theoremstyle{plain}
\newtheorem{lemma}{Lemma}
\theoremstyle{definition}

\input{macros.sty}
\input{physics_patch.sty}
\usepackage{tikz}
\usepackage{pgfplots}
\usepackage{pgfplotstable}
\usepackage{wrapfig}
\usepackage{subcaption}
\pgfplotsset{compat=1.18}

\iclrfinalcopy

\begin{document}

\maketitle
\begin{abstract}
State-of-the-art methods for forecasting irregularly sampled time series
with missing values predominantly rely on just four datasets and a few small toy examples for evaluation.
While ordinary differential equations (ODE) are the prevalent models in science and engineering, a baseline model that forecasts a constant value outperforms
ODE-based models from the last five years on three of these existing datasets.
This unintuitive finding hampers further research on ODE-based models, a more plausible model
family. In this paper, we develop a methodology to generate irregularly sampled
multivariate time series (IMTS) datasets from ordinary differential equations
and to select challenging instances via rejection sampling. Using this
methodology, we create~\Bench, a large and sophisticated benchmark of IMTS
datasets consisting of 50 individual datasets, derived from ODE models developed by 
research in Biology. \Bench~is the first
benchmark for IMTS forecasting that we are aware of and an order of magnitude larger
than the current evaluation setting. 
Using \Bench, we show qualitatively completely different results than those derived
from the current four datasets: on \Bench~deep learning methods based on ODEs can play to their
strength and our benchmark can differentiate in a meaningful way between
different IMTS forecasting models. This way, we expect to give a new impulse to research
on irregular time series modeling.

\end{abstract}

\section{Introduction}\label{sec:intro}
Over the past decade, substantial research has been devoted to fully observed
and regularly sampled time series, often referred to as regular (multivariate)
time series.
However, in certain scenarios, observations occur at irregular intervals
and variables are measured independently resulting in a sparse multivariate time series
with many missing values.
The resulting time series are known as irregularly sampled multivariate time series with missing values, or simply \textbf{IMTS}.

With regular time series forecasting being a well-established
topic, there exist a variety of sophisticated methods and standardized
benchmarks~\citep{Godahewa2021.Monasha, Gilpin2021.Chaosa,Bauer2021.Libra}.
On the other hand, there is currently no IMTS forecasting benchmark published.
Creating such a benchmark demands a high number of publicly available
time series datasets with genuinely irregular observation processes.
However, only a few of those exist. Although they are closely related
to each other, most evaluation sections of IMTS forecasting papers~\citep{Bilos2021.Neurald, Schirmer2022.Modelingb, Yalavarthi2024.GraFITi, Zhang2024.Irregular}
are based on at least a subset of these datasets.
Hence, it stands to reason that they induce a dataset bias.
Additionally, we show that a simple baseline, which always predicts a constant value independent of time, is competitive with or even outperforms complex neural ODE models on these datasets.
Hence, it appears questionable whether the currently used datasets are indeed well-suited for forecasting.


To address this limitation, we introduce \Bench, a wide benchmark of IMTS
datasets consisting of 50 individual datasets, derived from ordinary
differential equations from biological research, that are stored in the Physiome Model Repository (PMR).
Biological processes are well-suited for generating IMTS datasets, as they are inherently multivariate and irregularly measured in real-world experiments.
Additionally, the PMR provides Python implementations of many of these models, allowing us to create \Bench~in an automated manner.
While Biology researchers create their models based on very few and non-published observations,
they enable us to create an arbitrary number of time series which relate to possible measurements of a real-world phenomenon. 
\Bench~is the first benchmark for IMTS forecasting that we are aware of and an order of magnitude larger
than the current evaluation setting of just four datasets. 

Furthermore, to evaluate the complexity of the different forecasting datasets,
we introduce a simple metric called \emph{Joint Gradient Deviation ($\JGD$)}, which
measures the gradient variance of ODE solutions. We will show that
our benchmark consists of datasets of different complexity and covers a wide
range of different \emph{$\JGD$} values.

Finally, we evaluate current IMTS forecasting methods on $\Bench$ and
show that it includes many datasets on which neural ODE-based models significantly outperform
the time-constant baseline model. Furthermore, a member of the neural ODE
model family actually emerges as the overall most accurate model on \Bench.
However, the datasets in \Bench~are diverse enough that no single model is the most accurate for every dataset.

$\Bench$ is a significant step forward to standardized IMTS forecasting and to
monitor research progress.
Our contributions include the following:
\begin{enumerate}
\item We introduce a simple baseline model that is restricted to making
constant forecasts. On traditional IMTS forecasting
evaluation, this simple
baseline shows competitive or even better forecasting accuracy
when compared to ODE-based models.
\item We propose  \emph{Joint Gradient Deviation ($\JGD$)}, a simple score
designed to approximate the difficulty of
time series datasets for forecasting.
\item We create $\Bench$, a large and diverse benchmark of IMTS forecasting datasets.
Maximizing $\JGD$, we select and configure ODE models present in the Physiome~\citep{Yu2011.Physiome} repository.
$\Bench$ is the first benchmark for IMTS forecasting that we are aware of.
\item We evaluate state-of-the-art IMTS forecasting models on $\Bench$. 
Our experiments show that the datasets included in \Bench~are diverse enough to highlight different strengths of competing models. 
This results in different performance rankings of competing models for each dataset and the absence of a single best model. 
In that, \Bench~differs from the currently used IMTS forecasting datasets, which is expected to provide a fresh impulse to research
on IMTS forecasting models.
\end{enumerate}
We share our code on GitHub: \url{https://github.com/kloetergensc/Physiome-ODE}

\section{Irregularly Sampled Time Series Forecasting}\label{sec:prob_form}
%
An \emph{irregularly sampled multivariate time series (IMTS)} is a
sequence
  $X\coloneqq ((t_i, v_i))_{i=1:I} \in (\R \times (\R\cup \{\nan\})^C)^* \eqqcolon \mathcal{X}$
of pairs where
  $t_i$ is the observation timepoint and
  $v_i$ is an observation event
    with maximum of $C$-many variables (or channels) being observed.
We use $^*$ to indicate the space of sequences.
We call
  $v_{i,c} \in\R$ an observed value and
  $v_{i,c} = \nan$ a missing value.
$Q\coloneqq (t^\qry_k, c^\qry_k)_{k=1:K} \in (\R\times \{1, \ldots, C\})^*\eqqcolon \mathcal{Q}$
is the query sequence where
  $t^\qry_k$ is a queried (future) timepoint and
  $c^\qry_k$ is a queried channel.
$Y \coloneqq (y_1, \ldots, y_K) \in \R^* \eqqcolon \mathcal{Y}$
is the forecasting answer where
  $y_k$ is the ground truth forecast value for $Q_k$.

Then the \emph{IMTS forecasting problem} is defined as
follows:
given a sequence of triples $((X_n, Q_n, Y_n))_{n=1:N}$
drawn from a random distribution $\rho$ where
  $X_n \in\mathcal{X}$ is an irregularly sampled time series,
  $Q_n\in\mathcal{Q}$ is a query sequence and
  $Y_n \in \mathcal{Y}$ is a ground truth answer,
as well as
a loss function
  $\ell: \mathcal{Y} \times \mathcal{Y}\to \R$,
    e.g., the mean squared error between two equilength sequences,
the task is to find a model
  $\hat{Y}: \mathcal{X}\times \mathcal{Q} \to\mathcal{Y}$
such that the expected loss between
  the ground-truth forecasting answer $Y$ and
  the model prediction $\hat{Y}(X, Q)$ is minimal:
\begin{align}
	\mathcal{L}(\hat{Y}; \rho) \coloneqq \E_{(X, Q, Y)\sim \rho}
	\ell(Y, \hat{Y}(X,Q)).
\end{align}

\subsection{Current Forecasting Models}
In our experiments we compare five recent models.
Four of these five network architecture are variants of the neural
ODE~\citep{Chen2018.Neurald}, which uses an ODE, that is defined by a neural network, to infer the hidden state in-between observations. At
observation times the
neural ODE-based estimation of the hidden state is updated on the actual observation.
For more details, we refer to the papers. We use the following models:

\textbf{GRU-ODE-Bayes}~\citep{DeBrouwer2019.GRUODEBayesd}
utilizes a continuous version of Gated Recurrent Units (GRU) as
the ODE defining network $f$.
Additionally, GRU-Bayes applies a GRU-based Bayesian update to the
hidden state at observations.
\textbf{LinODEnet}~\citep{Scholz2022.Latenta} restricts $f$ to be linear functions but encodes data non-linearly into latent space.
 At observations LinODEnet updates its hidden state with a so-called nonlinear KalmanCell, a module inspired by
Kalman Filtering~\citep{Kalman1960.New}.
\textbf{Continuous Recurrent Units}~\citep{Schirmer2022.Modelingb} replaces the neural ODE with
 a stochastic differential equation (SDE). CRU is able to infer the hidden state
 at any time in closed form using continuous-discrete Kalman filtering,
 instead of solving a neural ODE\@.
\textbf{Neural Flows}~\citep{Bilos2021.Neurald} infers the solution curve of an ODE directly with
invertible neural networks.
\textbf{GraFITi}~\citep{Yalavarthi2024.GraFITi} is substantially different from neural ODE\@. It
encodes the observations from IMTS into a graph structure and infers forecasts by solving a
graph completion problem with a graph attention network. On the established evaluation datasets
GraFITi significantly outperforms the neural ODE-based models in terms of forecasting accuracy.

\section{Time-Constant Approaches Outperform ODE-based Approaches
in the Current Evaluation Scenario}\label{sec:classif_forec}

\paragraph{Current Evaluation Scenario.}
Traditionally, IMTS forecasting has relied on datasets like PhysioNet2012~\citep{Silva2012.Predictingb},
MIMIC-III~\citep{Johnson2016.MIMICIIIb}, and MIMIC-IV~\citep{Johnson.MIMICIV}
obtained from Kaggle challenges designed for IMTS classification.
Additionally, \citet{DeBrouwer2019.GRUODEBayesd} proposed to create an IMTS forecasting task,
based on USHCN~\citep{Menne2016.LongTerm}, a fully observed regularly sampled time series dataset
containing climate data. We present characteristics of these datasets in \Cref{tab:data} (appendix).
\begin{table}[t]
\caption{%
    Test MSE for forecasting next 50\% after 50\% observation time.
    OOM refers to out of memory. We highlight the best model in \textbf{bold}
	and \UL{underline} the second best. $^\dagger$ indicates that we show
	the results from \citet{Yalavarthi2024.GraFITi}
}\label{tab:classif_forec}
\centering
\begin{tabular}{l cccc}
	\toprule
	Model		& USHCN			& PhysioNet-2012	& MIMIC-III	& MIMIC-IV
\\	\midrule
	GRU-ODE     &  1.017±0.325  	& 0.653±0.023$^\dagger$       	& 0.653±0.023$^\dagger$      	& 0.439±0.003$^\dagger$
\\	LinODEnet   & \UL{0.662±0.126} 	& 0.411±0.001$^\dagger$  		& \UL{0.531 ± 0.022}$^\dagger$ 	& 0.336±0.002$^\dagger$
\\	CRU         & 0.730±0.264 		& 0.467±0.002$^\dagger$ 		& 0.619±0.028$^\dagger$         & OOM$^\dagger$
\\	Neural Flow & 1.014±0.336       & 0.506±0.002$^\dagger$        	& 0.651±0.017$^\dagger$         & 0.465±0.003$^\dagger$
\\	GraFITi     & \BF{0.636±0.161} 	& \BF{0.401±0.001}$^\dagger$  	& \BF{0.491 ± 0.014}$^\dagger$ 	& \BF{0.285±0.002}$^\dagger$
\\	GraFITi-C   & 0.875±0.204 		& \UL{0.407±0.001}				& 0.543±0.024 					& \UL{0.324±0.002}
\\ \bottomrule
\end{tabular}
\end{table}

The GraFITi model is the state-of-the-art on PhysioNet2012, MIMIC-III, MIMIC-IV, and USHCN, demonstrating a significant performance advantage over neural ODE-based methods~\citep{Yalavarthi2024.GraFITi}.
To further analyze this gap, we designed a \emph{constant}
version of GraFITi (GraFITi-C),
which is restricted to  make the exact same forecast
for every query time point.
This is achieved by training a GraFITi model that cannot access the actual query time point
($t^\qry_k$), but is instead always provided with a constant dummy query time point.

The results shown in \Cref{tab:classif_forec} indicate,
that neural ODE-based models are indeed outperformed
by this allegedly easy-to-beat baseline on PhysioNet2012 and
MIMIC-IV\@. On MIMIC-III the best performing ODE-based
model achieves a lower average test MSE than GraFITi-C,
however the difference is within the standard deviation.
GraFITi on the other hand, consistently outperforms its
constant variant. However, it also cannot separate in by a drastic margin.
These findings could relate to important covariants, which are not contained
in the dataset. For instance, it is impossible to predict a patient's blood glucose level without information about any food intake.

We designed \Bench~to be an extension to the currently used evaluation protocol.
For our ODE generated datasets, we can guarantee that all necessary covariants are given.
Furthermore, we can take full control about the simulated observation process.



\section{Generating Challenging IMTS Datasets from ODEs}

\paragraph{Sampling ODEs to Generate IMTS Data.}\label{sec:sampling}

In science and engineering, processes that evolve in continuous time are usually described by
differential equations, in the simpler cases by ordinary differential equations (ODEs).
All possible observations of such a process with $C$ \textbf{channels}
over \textbf{duration} $T$ at any time $t$ can be captured by a time-varying function
${x\colon [0,T]\to\R^C}$ as $x(t)_c$.
An ODE characterizes such a function implicitly by two conditions:
\begin{enumerate}
\item A relation between observation time $t$, observation value $x(t)$ and
the instantaneous change in observation values $x'(t)$, i.e.\ the first derivative of $x$.
This relation is described by a function $F: [0, T]\times\R^C\times\R^C\to\R$
called the \textbf{system}.

\item The \textbf{initial values} $x_0\in\R^C$ of the function at time $t=0$:
\begin{align}\label{eq:initial_value_problem}
	F(t, x(t), x'(t)) = 0 \quad \forall t\in[0,T], \qquad  x(0) = x_0
\end{align}
\end{enumerate}

Note that ODEs can be defined including higher order derivatives, but w.l.o.g.\ one can assume they are of first order, since the higher order system can be transformed into a first order system by a change of variables~\citep{Teschl2012.Ordinary}.
In a few special cases the solution $x(t)$ can be computed analytically. For example, the one-dimensional system $F(t, x(t), x'(t))\coloneqq a x(t) - x'(t)$ characterizes
the exponential growth function $x(t) = e^{at}x_0$. However, in most cases, it
can only be computed by a numerical ODE solver.

Specific parameters that occur in a system $F$ like the growth factor $a$
in our example are called \textbf{constants} and one writes $F(t, x(t), x'(t); a)$
with $a\in\R^A$ to denote a system with $A$-many constants.
Triples $(F, a, x_0)$ of systems, constants and initial values can be turned into a simulator,
i.e.\ a generative model, by complementing them with two more components:
\begin{enumerate}
\item A \textbf{sampling process} $s$ for observation time points and channels that
	can generate finite sequences $((t_1, c_1), \ldots, (t_I, c_I))$ of
	pairs of observation times $t_i\in[0,T]$ and observed channels $c_i\in\{1,\ldots,C\}$.
\item A \textbf{noise model} $p_\noise$ for the observation values, that is a conditional
	distribution ${p_\noise(x^\obs \mid x^\true)}$ of the observed value
	given the ground truth value.
\end{enumerate}
Then one can generate irregular sampled instances $(t_i, c_i, x^\obs_i)_{i=1{:}I}$
of the (only implicitly given) underlying function $x$ with missing values
and affected by observation noise simply via:
\begin{align}\label{eq:sampling}
		\bigl((t_i, c_i)\bigr)_{i=1:I} &\sim s
	&	x_i^\true  &\coloneqq \text{ode-solve}(F, a, x_0, t_i)_{c_i}
	&	x_i^\obs   &\sim p_\noise(x^\obs_i \mid x^\true_i)
\end{align}
%
%

\paragraph{Identifying Challenging Time Series.}\label{sec:mean-gradient-deviation}

While any ODE implicitly describes a time-varying function $x\colon [0,T] \to \R$,
not all such functions are challenging to forecast: For example,
some just describe very simple convergence processes that approach a limit value over a long time.
In order to discover challenging time series, we propose to measure the \emph{mean gradient deviation} ($\MGD$), that is the time-normalized $L^2$ norm of $x'(t)$ around its mean:
\begin{align}\label{eq:mdg}
	\MGD(x) &\coloneqq \min_{c} \sqrt{\frac{1}{T} \int_{0}^T (x'(t) - c)^2 \dd{t}}
\end{align}
Note that the minimum is obtained precisely for the mean gradient
\begin{equation}\label{eq:mgd}
	{c=\frac{1}{T}\int_{0}^T x'(t) \dd{t}} = \frac{x(T)-x(0)}{T}.
\end{equation}
For a linear test function $x(t) = a t + b$, the $\MGD$ is zero, while for a sine wave $x(t) = \sin(\omega t)$, the $\MGD$  scales linearly with the frequency $\omega$.
To identify not just individually challenging time series, but whole distributions (i.e.\ a generator $p$), one can compute the expectation of $\MGD$.
We found that the MGD grows super-exponentially with the Lipschitz constant, which is described in more detail in~\Cref{app:lip}

With MGD one does assess only the heterogeneity within each function,
not heterogeneity between different functions,
which is an important precondition for an interesting data set
for machine learning tasks. To capture this second aspect,
we propose the \emph{mean point-wise gradient deviation} ($\MPGD$):
\begin{align}\label{eq:mpgd}
	\MPGD(p) \coloneqq \frac{1}{T}\int_0^T  \std_{x\sim p}[x'(t)] \dd{t}
	= \frac{1}{T}\int_0^T\!\!\sqrt{%
		\E_{x\sim p}\bigl[\bigl(x'(t) - \E_{y\sim p}[y'(t)]\bigr)^2\bigr]
	}\dd{t}
\end{align}
We combine both measures multiplicatively as \emph{joint gradient deviation} ($\JGD$) for
distributions of functions, i.e.\ function generators:
\begin{align}\label{eq:jgd}
   \JGD(p) \coloneqq \MPGD(p) \cdot \E_{x\sim p}[\MGD(x)]
\end{align}
Note, that we can approximate $\MGD$ with a finite number of
samples:

\begin{lemma}\label{lemma:approx-tvl}
For a function $x \in \mathcal{C}^{1}([0,T])$ and $\epsilon > 0$ such that
$\frac{T}{\epsilon} \in \mathbb{N}$ we consider the divided differences
\begin{align*}
	x^{\diff,\epsilon}_t
	&\coloneqq \frac{x(t) - x(t-\epsilon)}{\epsilon}
	,\quad t=\epsilon,2\epsilon,\ldots,T.
\end{align*}
It then holds that the numerical estimator ${\numstd[(x_k)_{k=1:K}] \coloneqq \sqrt{\frac{1}{K}\sum_{k=1}^{K} (x_k - \bar{x})^2 }}$ of the standard deviation of the divided differences converges to the $\MGD$ of the function:
\begin{align}\label{eq:mgd_convergence}
	\numstd[x^{\diff,\epsilon}] \overset{\epsilon \to 0}{\longrightarrow} \MGD(x)
\end{align}
\end{lemma}

We can also approximate $\MPGD$ with a finite amount of samples under assumptions we explain in~\Cref{sec:proofs}.
\begin{lemma}\label{lemma:approx-atv}
The mean point-wise gradient deviation of a distribution $p$ of
functions can be approximated on a sample $(x_n)_{n=1:N}$ of $N$-many sequence from $p$ on a fixed grid of time points as follows: Given the divided differences:
\begin{align*}
	x^{\diff}_{n,t} & \coloneqq \frac{x_n(t) - x_n(t-\epsilon)}{\epsilon},\quad t=\epsilon,2\epsilon,\ldots,T
\end{align*}
Then the average of the point-wise std.\ of the divided differences converges to the $\MPGD$:
\begin{align}\label{eq:mpgd-approx}
\frac{\epsilon}{T}\sum_{t=\epsilon}^T \numstd[(x^{\text{diff}}_{n,t})_{n=1:N}]
	&\longrightarrow \MPGD(p)
	\qq{for} \epsilon\to 0 \qq{and} N\to\infty \qq{almost surely}
\end{align}
\end{lemma}
The proofs can be found in~\Cref{sec:proofs}. Due to \Cref{lemma:approx-tvl} and \Cref{lemma:approx-atv},
we can approximate the $\MGD$, $\MPGD$ and therefore the $\JGD$ based on the set of evenly spaced sequences ${X \in \R^{N \times T}}$, where $x_{n,t}$ denotes the $value$ at time-step $t$ in the $n$-th sequence, with $x_n \sim p$.
Given ${x^{\diff}_{n,t} \coloneqq  x_{n, t+1} - x_{n,t}}$,
it holds that ${\JGD(p) \approx  \MPGD(p) \cdot \frac{1}{N}\sum_{n=1}^{N} \MGD(x_n)}$ and furthermore:
\begin{align}\label{eq:jgd-approx}
	\MGD(x_n) &\approx \numstd[x^{\diff}_n]
&	\MPGD(p)  &\approx \frac{\epsilon}{T}\sum_{t=\epsilon}^{T}
	\numstd[(x^{\diff}_{n,t})_{n=1:N}]
\end{align}
\Cref{lemma:approx-tvl} and \Cref{lemma:approx-atv} thus allow us to approximate the $\JGD$-score by sampling the values of several functions on a shared grid from a generator.
Its definition on the full function guarantees that we will always get the same results (in the bounds of the approximation error).

\paragraph{JGD for ODE models.}\label{sec:jdg-ode-models}

To simplify the initial definition, we considered the $\JGD$ for a univariate function
${x\colon [0,T] \to \R}$, and we interpret a single channel within a multivariate ODE system as such a function.
Computing the $\JGD$ for a multivariate ODE model requires combining the $\JGD$ values of each channel.
We found that simply averaging all channels disadvantages models with a high number of channels.
To address this, we compute the mean of the ten channels with the highest $\JGD$\@.
For models with ten or fewer channels, we revert to using the mean of all channels.
Note that $\JGD$ is only a useful measure when applied to standardized data, otherwise the units are incomparable.
Consequently, we normalize each channel to have a mean of 0 and standard deviation of 1 over all
time steps and time series instances.

\paragraph{Varying Single ODE Instances.}\label{sec:pointwise-gradient-deviation}

From the scientific literature one often gets single ODE instances:
a triplet $(F, a^\lit, x_0^\lit)$ of a parametrized ODE system $F$,
its parameters/constants $a^\lit$ and the initial conditions $x_0^\lit$,
all three describing some specific experiment.
To generate data by sampling only from such a single ODE instance will create functions
that differ only in observation time points and channels and in the noise. These
instances are too similar to each other, without noise there is only one function,
and hence the $\MPGD$ of the distribution would be zero.
We therefore propose to also carefully vary
  (i) the initial conditions $x_0$,
  (ii) the ODE constants $a$ and
  (iii) the total duration $T$ and
of the ODE sampling process.
We sample each from a distribution 
that yields the values given in the scientific literature on expectation
and is controlled by a spread parameter $\sigma$ each:
\begin{align}\label{eq:ode-sampling}
	x_0	&\sim p_{\text{initial}}(\,\cdot\mid x_0^\lit, \sigma_{\text{initial}})
&	a	&\sim p_{\text{const}}(\,\cdot\mid a^\lit, \sigma_{\text{const}})
&	T	&\sim p_{\text{dur}}(\,\cdot\mid \sigma_{\text{dur}})
\end{align}
Let $p$ be the two-stage function generator that first samples tuples ${(x_0, a, T)}$
from their respective distributions and then yields as function $x$ the solution to the ODE
$F$ with initial values $x_0$, constants $a$ and running for a duration of $T$.
Instead of blindly choosing the spreads $\sigma$, we optimize them w.r.t.\ the $\JGD$-value of the generator:
\begin{align}\label{eq:optimal-sigma}
	(\sigma^*_{\text{initial}}, \sigma^*_{\text{const}}, \sigma^*_{\text{dur}})
	&\coloneqq \argmax_{\sigma_{\text{initial}}, \sigma_{\text{const}}, \sigma_{\text{dur}}}
	\JGD\bigl(
		p(\,\cdot \mid \sigma_{\text{initial}}, \sigma_{\text{const}}, \sigma_{\text{dur}})
	\bigr)
\end{align}
The $\JGD$ of each generator is estimated by 100 randomly sampled fully observed time series
with a sequence length of 100. Since we are interested in the difficulty of
the forecasting task, we compute the $\JGD$ on the final 50 steps of sequence only.
We search $\sigma_{\text{initial}}$ from the range of $\{0.1,0.3,0.5\}$,
$\sigma_{\text{const}}$ from $\{0.05,0.1,0.3\}$,
and $\sigma_{\text{dur}}$ from $\{0.33, 1, 3.3, 10, 30\}$.
The unit of $\sigma_\text{dur}$ varies for each ODE model,
it is given by Physiome~\citep{Yu2011.Physiome} and listed in \Cref{tab:ds-info} (appendix).

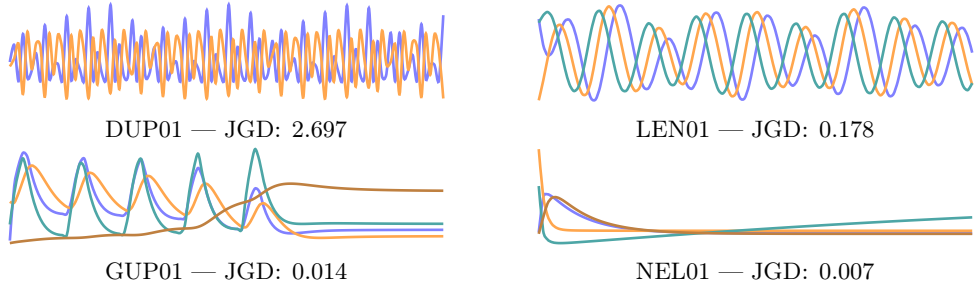
\begin{figure}
	\centering%
	\begin{subfigure}{0.495\textwidth}%
		\centering%
		\pgfplotstableread[col sep = comma]{figures/norm_dupont_1991a.csv}\tsa
\begin{tikzpicture}
\begin{axis}[
	width=\linewidth,
	height=1.5cm,
	scale only axis,
	ticks=none,
	hide axis,
	]
	\addplot[line width=1pt, smooth, color=blue!50] table[x=t, y=Z] {\tsa};
	\addplot[line width=1pt, color=orange!70, smooth] table[x=t, y=Y] {\tsa};
\end{axis}
\end{tikzpicture}%
		\caption*{DUP01 --- $\JGD$: 2.697}%
	\end{subfigure}
	\begin{subfigure}{0.495\textwidth}%
		\centering%
		\pgfplotstableread[col sep = comma]{figures/norm_lenbury_pacheenburawana_1991.csv}\tsb
\begin{tikzpicture}
	\begin{axis}[
		width=\linewidth,
		height=1.5cm,
		scale only axis,
		ticks=none,
		hide axis,
		]
		\addplot[line width=1pt, smooth, color=blue!50] table[x=t, y=Z] {\tsb};
		\addplot[line width=1pt, smooth, color=orange!70] table[x=t, y=Y] {\tsb};
		\addplot[line width=1pt, smooth, color=teal!70] table[x=t, y=X] {\tsb};
	\end{axis}
\end{tikzpicture}%
		\caption*{LEN01 --- $\JGD$: 0.178}
	\end{subfigure}

	\begin{subfigure}{0.495\linewidth}%
		\centering%
		\pgfplotstableread[col sep = comma]{figures/norm_gupta_aslakson_gurbaxani_vernon_2007_b.csv}\tsc
\begin{tikzpicture}
	\begin{axis}[
		width=\linewidth,
		height=1.5cm,
		scale only axis,
		ticks=none,
		hide axis,
		]
		\addplot[line width=1pt, smooth, color=blue!50] table[x=t, y=Z] {\tsc};
		\addplot[line width=1pt, smooth, color=orange!70] table[x=t, y=Y] {\tsc};
		\addplot[line width=1pt, smooth, color=teal!70] table[x=t, y=X] {\tsc};
		\addplot[line width=1pt, smooth, color=brown] table[x=t, y=W] {\tsc};
	\end{axis}
\end{tikzpicture}%
		\caption*{GUP01 --- $\JGD$: 0.014}
	\end{subfigure}
	\begin{subfigure}{0.495\linewidth}%
		\centering%
		\pgfplotstableread[col sep = comma]{figures/norm_nelson_murray_perelson_2000_general.csv}\tsd
\begin{tikzpicture}
\begin{axis}[
	width=\linewidth,
	height=1.5cm,
	scale only axis,
	ticks=none,
	hide axis,
	]
		\addplot[line width=1pt, smooth, color=blue!50] table[x=t, y=Z] {\tsd};
\addplot[line width=1pt, smooth, color=orange!70] table[x=t, y=Y] {\tsd};
\addplot[line width=1pt, smooth, color=teal!70] table[x=t, y=X] {\tsd};
\addplot[line width=1pt, smooth, color=brown] table[x=t, y=W] {\tsd};
\end{axis}
\end{tikzpicture}%
		\caption*{NEL01 --- $\JGD$: 0.007}
	\end{subfigure}
	\caption{Demonstration of time series realized by 4 ODEs of different prediction difficulties without adding any noise. Each line / color represents a channel. 
	Trajectories are shown for a duration of the respective $\sigma_\text{dur}$ as shown in \Cref{tab:ds-info}.}\label{fig:plots}
\end{figure}

\paragraph{Creating the $\Bench$ Benchmark.}\label{sec:rejection}
In the Physiome database we identified $N=208$ multivariate ODE models
${(F^n,a^{\lit,n},x_0^{\lit,n})_{n=1:N}}$, that are defined by
automatically generated Python code.
We ranked all ODE-models with according to their optimized $\JGD$-value
(see \Cref{tab:ds-info}) and selected the highest 50 datasets for
$\Bench$.
For each of the final datasets, we created a dataset with 2000 instances each.
To further increase the range of initial states, we create datasets of lengths
${\sigma^*_{\text{dur}}}$, but with 200 observation steps.
For each time series instance, we sample the initial time step
${t_\text{onset} \sim \{t_i\}_{1\leq i \leq 100}}$
and then select the next 100 steps for each time series instance.
Consequently, each time series instance has a duration of ${\frac{1}{2}\sigma^*_\text{dur}}$.
While the time series instances are sampled regularly at first,
we create IMTS instances by randomly dropping observations with a chance of 80\%.
For $p_{noise}$ we select to add $\epsilon \sim \mathcal{N} (0,0.05)$.
In \Cref{fig:plots} we show four trajectories contained in $\Bench$ with the
respective $\JGD$.

\paragraph{Handling exploding ODEs.}\label{sec:explosions}

Varying the constants and initial states of the ODEs can lead to inconsistent states,
which could never occur in real world scenarios.
In general, this is not an issue regarding the machine learning perspective on our experiments.
Occasionally however, certain combinations of inconsistent states and constants can lead to extreme values which are magnitudes greater than values that occur in other time series instances.
These extreme values will dominate the training and evaluation loss in a problematic manner.
We prevent this by rejecting all $(\sigma_\text{initial},\sigma_\text{const},\sigma_\text{dur})$ containing any values greater than 10 times the channel-wise standard deviation in the 100 samples created for evaluation.
If such values nevertheless occur within the final 2000 samples created for \Bench, we drop the respective instances.
We follow the same protocol to handle errors thrown by the ode-solver.



\section{Experiments}\label{sec:experiments}

\paragraph{Baseline models.}

We compare on 5 recent IMTS forecasting models, namely \textbf{GRU-ODE-Bayes}
\footnote{For brevity, we denote this as GRU-ODE in the tables.}~\citep{DeBrouwer2019.GRUODEBayesd}, \textbf{Neural Flows}~\citep{Bilos2021.Neurald}, Continuous Recurrent Unit (\textbf{CRU};~\citealp{Schirmer2022.Modelingb}), \textbf{LinODEnet}~\citep{Scholz2022.Latenta,mione_workflow_management_system_2024} and \textbf{GraFITi}~\citep{Yalavarthi2024.GraFITi}.
GRU-ODE-Bayes, Neural Flows, CRU and LinODEnet are differential equations based models whereas GraFITi is a graph based model.
We also introduce a naive model called GraFITi-Constant (GraFITi-C) which predicts a fixed value for all the future time points in all the target channels for an instance.

\paragraph{Experimental Protocol.}
In our experiments models have to predict the last 50\% of the time series after observing the first 50\%.
We use 5-fold cross validation, and for each fold we split the data into training, validation and test set with a ratio of 70:20:10.
Additionally, we resample the sparse observation mask to transform each instance into an IMTS\@.
The evaluation metric is the mean square error (MSE).
For simplicity, we randomly sample 10 hyperparameter configurations and fitted each model on a single fold per dataset, selecting the configuration with the lowest validation MSE\@.
The winning configuration per model is then trained and evaluated on all 5 folds (see \Cref{app:hype}).
We run the benchmark experiments on a cluster of 4 NVIDIA 4090 GPUs with 24 GB\@.

\paragraph{Empirical Results.}\label{sec:experiments_imts_bench}
The numbers presented in \Cref{tab:main_results} are the mean and standard deviation of MSE for 5 folds.
Overall, LinODEnet~\citep{Scholz2022.Latenta,mione_workflow_management_system_2024} is the best-performing model, winning the most datasets and having the highest average rank.
On the other hand, the average rank of GraFITi (2.20) is close to that of LinODEnet (2.10).
Surprisingly, GraFITi-C performs better than Neural Flows and GRU-ODE-Bayes and has an average rank of 2.86.
This is because datasets like \texttt{DUP01} and \texttt{DOK01} are too hard to forecast, hence the complex models cannot improve upon the accuracy of a constant model.
Other datasets are too simple that even a constant model yields MSE close to 0.0025,
which is the lowest value possible to achieve due to the normal-distributed noise with a variance of 0.05.

\begin{wrapfigure}{r}{0.45\textwidth}
	\centering
	\begin{subfigure}{\linewidth}
		\centering
	\pgfplotstableread[col sep = comma]{figures/lowest_error2.csv}\loadedtable
\begin{tikzpicture}
\begin{axis}[
	width=\linewidth,
	height=6cm,
	xmode=log,
	ymode=log,
	xlabel=$\JGD$,
	ylabel=Best MSE,
	]
	\addplot[only marks, scatter] table[x=high_score, y=lowest_mse] {\loadedtable};
\end{axis}
\end{tikzpicture}
	\end{subfigure}
	\caption{Test MSE of the best performing model vs $\JGD$-score across 50 datasets.}\label{fig:mse_jgd}
	\vspace{-\baselineskip}
\end{wrapfigure}
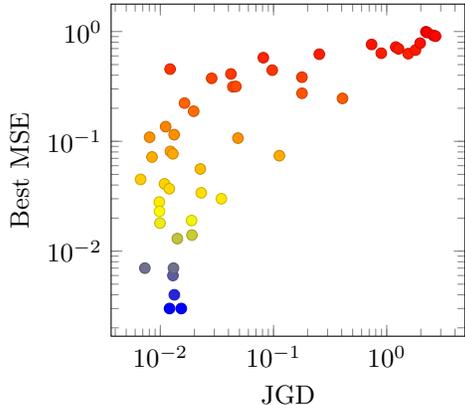

In order to check how good our $\JGD$ score can measure the difficulty in forecasting of a time series we plot $\JGD$ score vs the best MSE score achieved in \Cref{fig:mse_jgd}.
It can be observed that $\JGD$ fulfills its purpose of
finding challenging and interesting ODE models. More specifically $\JGD$ and the test MSE of the
best performing model are correlated with a Spearman coefficient of 0.77.
\begin{table}
\setlength{\tabcolsep}{1.6mm}
\caption{%
	Experimental results on various baseline models.
	\Bench~datasets ranked by $\JGD$-score.
	We highlight the best model in \textbf{bold}
	and \UL{underline} the second best.
}\label{tab:main_results}
\centering
\scalebox{1}{
\small
\begin{tabular}{lccccccc}
\toprule
Dataset & $\JGD$ & GRU-ODE & LinODEnet & CRU & Neural Flows & GraFITi & GraFITi-C \\	
\midrule
DUP01 & 2.697 &  1.037±0.047 &      0.964±0.036 &      0.958±0.038 & 1.030±0.046 & \UL{0.955±0.037} & \BF{0.951±0.036} \\
JEL01 & 2.580 &  1.017±0.014 &      0.949±0.015 & \UL{0.939±0.015} & 1.000±0.013 &      0.942±0.020 & \BF{0.935±0.016} \\
DOK01 & 2.277 &  1.011±0.004 &      0.996±0.003 &      0.985±0.005 & 0.998±0.005 & \UL{0.984±0.005} & \BF{0.982±0.005} \\
INA01 & 2.218 &  1.018±0.010 &      1.009±0.011 &      1.005±0.010 & 1.008±0.008 & \BF{1.004±0.010} & \BF{1.004±0.009} \\
WOL01 & 1.973 &  0.952±0.035 &      0.806±0.027 &      0.814±0.029 & 0.841±0.029 & \UL{0.787±0.028} & \BF{0.784±0.030} \\
BOR01 & 1.795 &  0.794±0.034 &      0.719±0.021 &      0.715±0.020 & 0.743±0.026 & \UL{0.712±0.022} & \BF{0.709±0.022} \\
HYN01 & 1.548 &  0.883±0.060 &      0.672±0.044 &      0.665±0.053 & 0.636±0.045 & \UL{0.625±0.043} & \BF{0.619±0.046} \\
JEL02 & 1.271 &  0.816±0.049 &      0.693±0.031 & \BF{0.674±0.028} & 0.779±0.028 &      0.699±0.029 & \UL{0.687±0.027} \\
DUP02 & 1.202 &  0.895±0.055 &      0.740±0.042 & \UL{0.722±0.046} & 0.890±0.081 &      0.728±0.044 & \BF{0.718±0.046} \\
WOL02 & 0.895 &  0.854±0.010 &      0.663±0.015 & \UL{0.653±0.017} & 0.685±0.012 &      0.654±0.014 & \BF{0.645±0.016} \\
DIF01 & 0.735 &  1.035±0.023 & \BF{0.832±0.087} &      0.985±0.025 & 1.014±0.025 &      0.985±0.030 & \UL{0.982±0.029} \\
VAN01 & 0.407 &  0.321±0.023 &      0.250±0.006 &      0.253±0.005 & 0.250±0.006 & \UL{0.246±0.005} & \BF{0.242±0.006} \\
DUP03 & 0.254 &  0.874±0.089 &      0.632±0.044 & \BF{0.622±0.047} & 1.098±0.447 & \UL{0.627±0.043} &      0.744±0.042 \\
BER01 & 0.179 &  0.594±0.054 & \BF{0.279±0.020} & \UL{0.280±0.016} & 0.398±0.014 &      0.300±0.018 &      0.342±0.018 \\
LEN01 & 0.178 &  1.028±0.060 & \BF{0.387±0.071} &      0.754±0.157 & 1.009±0.061 & \UL{0.607±0.055} &      0.970±0.063 \\
 LI01 & 0.113 &  1.067±0.015 & \BF{0.084±0.009} & \UL{0.175±0.020} & 0.979±0.049 &      0.202±0.013 &      0.742±0.010 \\
 LI02 & 0.097 &  0.723±0.080 & \UL{0.434±0.044} &      0.437±0.046 & 0.674±0.105 & \BF{0.397±0.058} &      0.458±0.056 \\
REV01 & 0.081 &  1.091±0.075 & \BF{0.597±0.061} & \UL{0.602±0.049} & 0.978±0.042 &      0.674±0.055 &      0.855±0.050 \\
PUR01 & 0.049 &  0.703±0.058 & \BF{0.106±0.006} &      0.353±0.083 & 0.654±0.102 & \UL{0.153±0.006} &      0.476±0.020 \\
NYG01 & 0.047 &  0.571±0.074 & \UL{0.358±0.071} &      0.403±0.092 & 0.442±0.094 & \BF{0.344±0.065} &      0.366±0.047 \\
PUR02 & 0.044 &  0.830±0.066 & \BF{0.280±0.028} & \UL{0.293±0.026} & 0.723±0.077 &      0.322±0.021 &      0.511±0.023 \\
HOD01 & 0.042 &  0.851±0.118 & \UL{0.441±0.043} & \BF{0.409±0.049} & 0.701±0.077 &      0.493±0.046 &      0.609±0.056 \\
REE01 & 0.035 &  0.266±0.068 &      0.045±0.012 &      0.051±0.008 & 0.045±0.011 & \BF{0.033±0.007} & \UL{0.039±0.012} \\
VIL01 & 0.028 &  0.511±0.053 &      0.374±0.021 & \UL{0.373±0.039} & 0.500±0.060 & \BF{0.344±0.044} &      0.378±0.042 \\
KAR01 & 0.023 &  0.193±0.013 & \BF{0.034±0.008} &      0.044±0.012 & 0.069±0.009 & \UL{0.041±0.013} &      0.078±0.011 \\
SHO01 & 0.023 &  0.260±0.017 & \UL{0.057±0.006} &      0.095±0.010 & 0.092±0.014 &      0.062±0.013 & \BF{0.055±0.013} \\
BUT01 & 0.020 &  0.583±0.172 & \BF{0.254±0.074} &      0.317±0.108 & 0.441±0.153 & \UL{0.281±0.071} &      0.324±0.091 \\
MAL01 & 0.019 &  0.420±0.061 & \BF{0.018±0.007} &      0.064±0.007 & 0.052±0.002 & \UL{0.020±0.004} &      0.054±0.005 \\
ASL01 & 0.019 &  0.114±0.035 & \BF{0.022±0.003} &      0.046±0.014 & 0.066±0.033 & \UL{0.025±0.009} &      0.026±0.002 \\
BUT02 & 0.016 &  0.483±0.047 & \BF{0.207±0.056} &      0.282±0.042 & 0.329±0.068 & \UL{0.248±0.052} &      0.256±0.039 \\
MIT01 & 0.015 &  0.005±0.001 & \BF{0.003±0.000} & \BF{0.003±0.000} & 0.008±0.004 & \BF{0.003±0.000} & \BF{0.003±0.000} \\
GUP01 & 0.014 &  0.153±0.041 & \BF{0.018±0.007} &      0.057±0.017 & 0.043±0.017 &      0.041±0.006 & \UL{0.035±0.006} \\
GUY01 & 0.013 &  0.036±0.004 &      0.006±0.005 & \BF{0.004±0.001} & 0.024±0.015 &      0.005±0.003 & \BF{0.004±0.001} \\
PHI01 & 0.013 &  0.674±0.030 & \BF{0.131±0.014} & \UL{0.133±0.020} & 0.635±0.158 &      0.222±0.013 &      0.345±0.015 \\
GUY02 & 0.013 &  0.124±0.044 & \BF{0.010±0.006} & \BF{0.010±0.002} & 0.082±0.066 &      0.012±0.009 &      0.032±0.015 \\
PUL01 & 0.013 &  0.099±0.024 & \BF{0.008±0.004} &      0.012±0.003 & 0.061±0.060 & \BF{0.008±0.001} &      0.024±0.008 \\
CAL01 & 0.013 &  1.049±0.055 & \BF{0.078±0.009} & \UL{0.158±0.008} & 0.867±0.014 &      0.179±0.012 &      0.643±0.024 \\
WOD01 & 0.012 &  0.612±0.072 & \UL{0.154±0.016} & \BF{0.113±0.017} & 0.510±0.060 &      0.164±0.013 &      0.344±0.016 \\
GUP02 & 0.012 &  0.870±0.059 &      0.469±0.022 & \BF{0.444±0.018} & 0.577±0.017 & \UL{0.449±0.027} &      0.461±0.025 \\
  M01 & 0.012 &  0.055±0.009 &      0.004±0.001 &      0.005±0.000 & 0.124±0.207 & \BF{0.003±0.000} & \BF{0.003±0.000} \\
LEN02 & 0.012 &  0.297±0.071 & \BF{0.039±0.005} & \UL{0.059±0.012} & 0.380±0.141 &      0.099±0.021 &      0.143±0.022 \\
KAR02 & 0.011 &  0.515±0.032 & \BF{0.140±0.010} & \UL{0.151±0.011} & 0.257±0.016 & \UL{0.151±0.009} &      0.252±0.010 \\
SHO02 & 0.011 &  0.368±0.028 & \BF{0.037±0.006} &      0.083±0.015 & 0.109±0.015 & \UL{0.043±0.006} &      0.073±0.010 \\
MAC01 & 0.010 &  0.242±0.026 & \UL{0.020±0.003} &      0.065±0.006 & 0.029±0.011 &      0.021±0.003 & \BF{0.019±0.002} \\
IRI01 & 0.010 &  0.151±0.032 & \BF{0.037±0.003} &      0.049±0.010 & 0.116±0.006 & \UL{0.038±0.017} &      0.097±0.008 \\
BAG01 & 0.010 &  0.294±0.041 & \UL{0.032±0.005} &      0.046±0.005 & 0.075±0.012 & \BF{0.029±0.002} &      0.109±0.002 \\
WOL03 & 0.008 &  0.859±0.114 & \BF{0.073±0.010} &      0.177±0.016 & 0.479±0.107 & \UL{0.105±0.016} &      0.247±0.032 \\
WAN01 & 0.008 &  0.504±0.031 & \BF{0.103±0.012} &      0.125±0.012 & 0.345±0.027 & \UL{0.119±0.010} &      0.232±0.015 \\
NEL01 & 0.007 &  0.054±0.012 &      0.010±0.001 & \UL{0.009±0.001} & 0.060±0.029 & \BF{0.007±0.000} &      0.023±0.006 \\
HUA01 & 0.007 &  0.321±0.046 & \BF{0.052±0.004} &      0.116±0.007 & 0.149±0.015 & \UL{0.063±0.005} &      0.115±0.007 \\
\midrule 
\# Wins	&	---	&	0	&	\BF{25}	&	8	&	2	&	10	&	 \UL{15}\\
Rank &	---	&	5.9	&	\BF{2.10}	&	2.82	&	4.82	&	\UL{2.20}	&	2.86 \\	
\bottomrule
\end{tabular}

}
\end{table}

\normalsize
\section{Extensions and Limitations}\label{sec:limit}
Currently, most datasets in \Bench~still show a good potential for improvement:
even the best performing methods are far away from the Bayes error of
0.0025 MSE\@.
In case authors of future methods get the impression that their method
could benefit from larger datasets, i.e., more instances per dataset than 2000,
they easily can generate larger ones with 10,000, 100,000, etc.\ instances.

\Bench~is not based on observation data as in these areas
observed data usually is expensive to get, and thus only limited
amounts exists or are publically available, at least.
In fact, we scanned through the respective papers of our top 50 ODE models 
and could not find the original experimental data for a single one. 
Hence, if one wants to conduct machine learning experiments on these complex
and rich biological processes one is forced to create them using the ODE models describing the process.

Consequently, we build the benchmark on top of real-world ODEs:
their system, their constants and their initial values establish
a real-world connection that has been created in hundreds
of scientific publications. To delineate our benchmark from purely
synthetic data we therefore call it \emph{semi-synthetic}.

The fact that the datasets are generated by ODEs implies that some general ODE-based models are expressive enough to mimic the generating process and to perform well for large amounts of data.
For example, LinODEnet can represent a Koopman-type~\citep{Koopman1931.Hamiltonian,Koopman1932.Dynamical} linear approximation of any nonlinear ODE under some assumptions. 
However, being expressive enough does not mean that these models are  best when trained on a limited amount of data.
Other, non-ODE models can beat ODE-based models with limited training data, even if the ground truth is an ODE\@.
In fact, a main purpose of the benchmark is to answer questions about the best forecasting models for data generated from an (unknown) ODE ground truth.  
Despite our limitation to ODEs from biology the perspective is forecasting for many scientific domains where dynamics are \emph{in principle} governed by ODEs.

The inductive bias of ODE-based models (LinODEnet~\citep{Scholz2022.Latenta}) is rather weak.
Being based on ODEs does not mean that these models would only perform well on data generated from ODEs.
When one spells out the expressiveness of LinODEnet in terms of basis functions,
it would just mean that after a nonlinear transformation the time dependent functions can be expressed by some basis functions given as solutions of some linear ODE\@. 
The representation is entirely trainable and nothing about the time series in question is hard-coded apart from smoothness.

The number of instances per dataset contained in \Bench~is deliberately small (2000). Our goal was not to create
huge datasets just for the sake of being huge, but to create
datasets for the purpose of research: the smallest datasets that
would allow complex, ODE-based models to demonstrate their
strengths. However, it is trivial to create an arbitrary number of time series, if larger datasets are desired.  

While this work focuses on IMTS forecasting, \Bench~can be easily extended to regular time series forecasting.
In \Cref{app:regular}, we conduct experiments with three datasets created with \Bench. 
Our results show that these datasets have an interesting property: They force models to learn channel dependencies. 
This contradicts with the results on traditional evaluation datasets, in which models like PatchTST~\citep{Nie2023.Time} are more effective when they ignore 
inter-channel correlations.  

\section{Prior Time Series Forecasting Benchmarks}\label{sec:related}
\paragraph{Monash time series forecasting archive~\citep{Godahewa2021.Monasha}}
The Monash repository is the first benchmark for time series forecasting.
It encompasses 11 openly-accessible multivariate time series datasets drawn from various
domains, including energy, banking, and nature. Some of these datasets may include
missing values. However, the sampling process is regular.
Therefore, the missing values can easily be imputed, so that methods designed for regular
time series forecasting can be applied.
Hypothetically, we could use the 11 multivariate time series contained in
the Monash archive to create IMTS datasets for an IMTS forecasting benchmark.
However, most of these multivariate datasets are actually multiple
univariate time series stacked on top of each other and/or a single
long time series that is segmented to create multiple time series instances.
Other datasets used to evaluate regular multivariate time series
forecasting~\citep{Nie2023.Time,Zeng2023.Are,Zhou2021.Informera},
are not well-suited for IMTS forecasting for similar reasons. In fact,
\citet{Nie2023.Time} showed that models benefit, when they ignore channel
correlations, hinting that these are barely carrying useful information in the
respective datasets. In IMTS forecasting however, modeling these channel correlations
is a crucial property of the models, due to the high number of missing values.

\paragraph{Chaotic ODEs benchmark for time series forecasting}
Closely related to \Bench, \citet{Gilpin2021.Chaosa} provides a forecasting benchmark
created with chaotic ordinary differential equations that mainly originate from Physics.
Similar to the Monash repository, each chaotic ODE creates single long trajectory, without
varying the constants. In contrast, \Bench~contains 2000 different trajectories with varying
constants and initial states.
Furthermore, the inherent chaotic nature of these equations often presents a forecasting
challenge, even in the regularly sampled and fully observed forecasting setting.
On the other hand, IMTS forecasting is challenging due to the sparse observation of a time series.
We assume that combining these two factors of difficulty, would surpass any current methods
forecast modeling ability. This is closely related to what we observe on the most difficult
dataset of \Bench. We support our assumption, by an example experiment
on the Lorenz-attractor, a famous chaotic ODE, which is a part of \citet{Gilpin2021.Chaosa}'s
benchmark. Respective experimental details are given in \Cref{app:lorenz} and results
in \Cref{tab:lorenz}. Similarly, to certain \Bench~datasets with a high $\JGD$
(e.g.\ DUP01), no method can significantly outperform the simple baseline GraFITi-C.

\paragraph{PDEBench~\citep{Takamoto2022.Pdebencha}}
Related to applying differential equations to create datasets, there exists a benchmark with 11 spatio-temporal datasets named PDEBench.
Here, models are trained to estimate the parameters of an underlying partial differential equation (PDE) used to generate such dataset, instead of forecasting.
Current IMTS forecasting architectures cannot utilize the spatio-temporal information inherent in PDEBench's datasets.

\section{Conclusion}
We introduced \Bench, the first ever IMTS forecasting benchmark.
\Bench's datasets are created with ODE models, that were defined in decades of research and published on
the Physiome Model Repository. Our experiments showed that LinODEnet and CRU are actually
better than previous evaluation on established datasets indicated. Nevertheless,
we also provided a few datasets, on which models are unable to outperform a
constant baseline model. We believe that our datasets, especially the very difficult ones,
can help to identify deficits of current architectures and support future research on
IMTS forecasting.

\section*{Reproducibility Statement}

Our experiments can be reproduced by following the instructions provided in our Git repository:
\url{https://github.com/kloetergensc/Physiome-ODE}. There are two options to obtain the datasets from \Bench. 
The first is to regenerate everything using the code and instructions in the repository. 
The second is to download the data from Zenodo \url{https://zenodo.org/records/11492058}:

\newpage

\bibliographystyle{abbrvnat}
\bibliography{references}

\begin{thebibliography}{27}
\providecommand{\natexlab}[1]{#1}
\providecommand{\url}[1]{\texttt{#1}}
\expandafter\ifx\csname urlstyle\endcsname\relax
  \providecommand{\doi}[1]{doi: #1}\else
  \providecommand{\doi}{doi: \begingroup \urlstyle{rm}\Url}\fi

\bibitem[Andrews(1987)]{andrews_consistency_1987}
D.~W.~K. Andrews.
\newblock Consistency in {Nonlinear} {Econometric} {Models}: {A} {Generic}
  {Uniform} {Law} of {Large} {Numbers}.
\newblock \emph{Econometrica}, 55\penalty0 (6):\penalty0 1465, Nov. 1987.
\newblock ISSN 00129682.
\newblock \doi{10.2307/1913568}.
\newblock URL \url{https://www.jstor.org/stable/1913568?origin=crossref}.

\bibitem[Bauer et~al.(2021)Bauer, Z{\"u}fle, Eismann, Grohmann, Herbst, and
  Kounev]{Bauer2021.Libra}
A.~Bauer, M.~Z{\"u}fle, S.~Eismann, J.~Grohmann, N.~Herbst, and S.~Kounev.
\newblock Libra: {{A Benchmark}} for {{Time Series Forecasting Methods}}.
\newblock In \emph{Proceedings of the {{ACM}}/{{SPEC International Conference}}
  on {{Performance Engineering}}}, {{ICPE}} '21, pages 189--200, New York, NY,
  USA, Apr. 2021. Association for Computing Machinery.
\newblock ISBN 978-1-4503-8194-9.
\newblock \doi{10.1145/3427921.3450241}.

\bibitem[Bilo{\v s} et~al.(2021)Bilo{\v s}, Sommer, Rangapuram, Januschowski,
  and G{\"u}nnemann]{Bilos2021.Neurald}
M.~Bilo{\v s}, J.~Sommer, S.~S. Rangapuram, T.~Januschowski, and
  S.~G{\"u}nnemann.
\newblock Neural {{Flows}}: {{Efficient Alternative}} to {{Neural ODEs}}.
\newblock In \emph{Advances in {{Neural Information Processing Systems}}},
  volume~34, pages 21325--21337. Curran Associates, Inc., 2021.

\bibitem[Chen et~al.(2018)Chen, Rubanova, Bettencourt, and
  Duvenaud]{Chen2018.Neurald}
R.~T.~Q. Chen, Y.~Rubanova, J.~Bettencourt, and D.~K. Duvenaud.
\newblock Neural {{Ordinary Differential Equations}}.
\newblock In \emph{Advances in {{Neural Information Processing Systems}}},
  volume~31. Curran Associates, Inc., 2018.

\bibitem[Chen et~al.(2023)Chen, Li, Yoder, Arik, and
  Pfister]{Chen2023.TSMixerb}
S.-A. Chen, C.-L. Li, N.~Yoder, S.~O. Arik, and T.~Pfister.
\newblock {{TSMixer}}: {{An All-MLP Architecture}} for {{Time Series
  Forecasting}}, Sept. 2023.

\bibitem[De~Brouwer et~al.(2019)De~Brouwer, Simm, Arany, and
  Moreau]{DeBrouwer2019.GRUODEBayesd}
E.~De~Brouwer, J.~Simm, A.~Arany, and Y.~Moreau.
\newblock {{GRU-ODE-Bayes}}: {{Continuous Modeling}} of {{Sporadically-Observed
  Time Series}}.
\newblock In \emph{Advances in {{Neural Information Processing Systems}}},
  volume~32. Curran Associates, Inc., 2019.

\bibitem[Gilpin(2021)]{Gilpin2021.Chaosa}
W.~Gilpin.
\newblock Chaos as an interpretable benchmark for forecasting and data-driven
  modelling.
\newblock In \emph{Thirty-Fifth {{Conference}} on {{Neural Information
  Processing Systems Datasets}} and {{Benchmarks Track}} ({{Round}} 2)}, Aug.
  2021.

\bibitem[Godahewa et~al.(2021)Godahewa, Bergmeir, Webb, Hyndman, and
  {Montero-Manso}]{Godahewa2021.Monasha}
R.~W. Godahewa, C.~Bergmeir, G.~Webb, R.~Hyndman, and P.~{Montero-Manso}.
\newblock Monash {{Time Series Forecasting Archive}}.
\newblock \emph{Proceedings of the Neural Information Processing Systems Track
  on Datasets and Benchmarks}, 1, Dec. 2021.

\bibitem[Jennrich(1969)]{jennrich_asymptotic_properties_non-linear_1969}
R.~I. Jennrich.
\newblock Asymptotic {{Properties}} of {{Non-Linear Least Squares Estimators}}.
\newblock \emph{The Annals of Mathematical Statistics}, 40\penalty0
  (2):\penalty0 633--643, 1969.
\newblock ISSN 0003-4851.
\newblock \doi{10.1214/aoms/1177697731}.

\bibitem[Johnson et~al.(2021)Johnson, Bulgarelli, Pollard, Horng, Celi, and
  Mark]{Johnson.MIMICIV}
A.~Johnson, L.~Bulgarelli, T.~Pollard, S.~Horng, L.~A. Celi, and R.~Mark.
\newblock {{MIMIC-IV}}.
\newblock \emph{PhysioNet}, 2021.
\newblock \doi{10.13026/RRGF-XW32}.

\bibitem[Johnson et~al.(2016)Johnson, Pollard, Shen, Lehman, Feng, Ghassemi,
  Moody, Szolovits, Anthony~Celi, and Mark]{Johnson2016.MIMICIIIb}
A.~E.~W. Johnson, T.~J. Pollard, L.~Shen, L.-w.~H. Lehman, M.~Feng,
  M.~Ghassemi, B.~Moody, P.~Szolovits, L.~Anthony~Celi, and R.~G. Mark.
\newblock {{MIMIC-III}}, a freely accessible critical care database.
\newblock \emph{Scientific Data}, 3\penalty0 (1):\penalty0 160035, May 2016.
\newblock ISSN 2052-4463.
\newblock \doi{10.1038/sdata.2016.35}.

\bibitem[Kalman(1960)]{Kalman1960.New}
R.~E. Kalman.
\newblock A {{New Approach}} to {{Linear Filtering}} and {{Prediction
  Problems}}.
\newblock \emph{Journal of Basic Engineering}, 82\penalty0 (1):\penalty0
  35--45, Mar. 1960.
\newblock ISSN 0021-9223.
\newblock \doi{10.1115/1.3662552}.

\bibitem[Koopman(1931)]{Koopman1931.Hamiltonian}
B.~O. Koopman.
\newblock Hamiltonian {{Systems}} and {{Transformation}} in {{Hilbert Space}}.
\newblock \emph{Proceedings of the National Academy of Sciences}, 17\penalty0
  (5):\penalty0 315--318, May 1931.
\newblock \doi{10.1073/pnas.17.5.315}.

\bibitem[Koopman and v.~Neumann(1932)]{Koopman1932.Dynamical}
B.~O. Koopman and J.~v.~Neumann.
\newblock Dynamical {{Systems}} of {{Continuous Spectra}}.
\newblock \emph{Proceedings of the National Academy of Sciences}, 18\penalty0
  (3):\penalty0 255--263, Mar. 1932.
\newblock \doi{10.1073/pnas.18.3.255}.

\bibitem[Menne et~al.(2016)Menne, Williams, and Vose]{Menne2016.LongTerm}
M.~J. Menne, J.~Williams, and R.~S. Vose.
\newblock Long-{{Term Daily}} and {{Monthly Climate Records}} from {{Stations
  Across}} the {{Contiguous United States}} ({{U}}.{{S}}. {{Historical
  Climatology Network}}).
\newblock Technical Report osti:1394920; cdiac:NDP-019;
  doi:10.3334/CDIAC/CLI.NDP019, Environmental System Science Data
  Infrastructure for a Virtual Ecosystem (ESS-DIVE) (United States); CDIAC,
  Jan. 2016.

\bibitem[Mione et~al.(2024)Mione, Kaspersetz, Luna, Aizpuru, Scholz, Borisyak,
  Kemmer, Schermeyer, Martinez, Neubauer, and
  Cruz~Bournazou]{mione_workflow_management_system_2024}
F.~M. Mione, L.~Kaspersetz, M.~F. Luna, J.~Aizpuru, R.~Scholz, M.~Borisyak,
  A.~Kemmer, M.~T. Schermeyer, E.~C. Martinez, P.~Neubauer, and M.~N.
  Cruz~Bournazou.
\newblock A workflow management system for reproducible and interoperable
  high-throughput self-driving experiments.
\newblock \emph{Computers \& Chemical Engineering}, 187:\penalty0 108720, Aug.
  2024.
\newblock ISSN 0098-1354.
\newblock \doi{10.1016/j.compchemeng.2024.108720}.

\bibitem[Nie et~al.(2023)Nie, Nguyen, Sinthong, and Kalagnanam]{Nie2023.Time}
Y.~Nie, N.~H. Nguyen, P.~Sinthong, and J.~Kalagnanam.
\newblock A {{Time Series}} is {{Worth}} 64 {{Words}}: {{Long-term
  Forecasting}} with {{Transformers}}.
\newblock In \emph{The {{Eleventh International Conference}} on {{Learning
  Representations}}, {{ICLR}} 2023, {{Kigali}}, {{Rwanda}}, {{May}} 1-5, 2023}.
  OpenReview.net, 2023.

\bibitem[Schirmer et~al.(2022)Schirmer, Eltayeb, Lessmann, and
  Rudolph]{Schirmer2022.Modelingb}
M.~Schirmer, M.~Eltayeb, S.~Lessmann, and M.~Rudolph.
\newblock Modeling {{Irregular Time Series}} with {{Continuous Recurrent
  Units}}.
\newblock In \emph{Proceedings of the 39th {{International Conference}} on
  {{Machine Learning}}}, pages 19388--19405. PMLR, June 2022.

\bibitem[Scholz et~al.(2022)Scholz, Born, {Duong-Trung}, {Cruz-Bournazou}, and
  {Schmidt-Thieme}]{Scholz2022.Latenta}
R.~Scholz, S.~Born, N.~{Duong-Trung}, M.~N. {Cruz-Bournazou}, and
  L.~{Schmidt-Thieme}.
\newblock Latent {{Linear ODEs}} with {{Neural Kalman Filtering}} for
  {{Irregular Time Series Forecasting}}.
\newblock Sept. 2022.

\bibitem[Silva et~al.(2012)Silva, Moody, Scott, Celi, and
  Mark]{Silva2012.Predictingb}
I.~Silva, G.~Moody, D.~J. Scott, L.~A. Celi, and R.~G. Mark.
\newblock Predicting in-hospital mortality of {{ICU}} patients: {{The
  PhysioNet}}/{{Computing}} in cardiology challenge 2012.
\newblock In \emph{2012 {{Computing}} in {{Cardiology}}}, pages 245--248, Sept.
  2012.

\bibitem[Takamoto et~al.(2022)Takamoto, Praditia, Leiteritz, MacKinlay,
  Alesiani, Pfl{\"u}ger, and Niepert]{Takamoto2022.Pdebencha}
M.~Takamoto, T.~Praditia, R.~Leiteritz, D.~MacKinlay, F.~Alesiani,
  D.~Pfl{\"u}ger, and M.~Niepert.
\newblock Pdebench: {{An}} extensive benchmark for scientific machine learning.
\newblock \emph{Advances in Neural Information Processing Systems},
  35:\penalty0 1596--1611, 2022.

\bibitem[Teschl(2012)]{Teschl2012.Ordinary}
G.~Teschl.
\newblock \emph{Ordinary {{Differential Equations}} and {{Dynamical Systems}}}.
\newblock American Mathematical Soc., Aug. 2012.
\newblock ISBN 978-0-8218-8328-0.

\bibitem[Yalavarthi et~al.(2024)Yalavarthi, Madhusudhanan, Scholz, Ahmed,
  Burchert, Jawed, Born, and {Schmidt-Thieme}]{Yalavarthi2024.GraFITi}
V.~K. Yalavarthi, K.~Madhusudhanan, R.~Scholz, N.~Ahmed, J.~Burchert, S.~Jawed,
  S.~Born, and L.~{Schmidt-Thieme}.
\newblock {{GraFITi}}: {{Graphs}} for {{Forecasting Irregularly Sampled Time
  Series}}.
\newblock In M.~J. Wooldridge, J.~G. Dy, and S.~Natarajan, editors,
  \emph{Thirty-{{Eighth AAAI Conference}} on {{Artificial Intelligence}},
  {{AAAI}} 2024, {{Thirty-Sixth Conference}} on {{Innovative Applications}} of
  {{Artificial Intelligence}}, {{IAAI}} 2024, {{Fourteenth Symposium}} on
  {{Educational Advances}} in {{Artificial Intelligence}}, {{EAAI}} 2014,
  {{February}} 20-27, 2024, {{Vancouver}}, {{Canada}}}, pages 16255--16263.
  AAAI Press, 2024.
\newblock \doi{10.1609/AAAI.V38I15.29560}.

\bibitem[Yu et~al.(2011)Yu, Lloyd, Nickerson, Cooling, Miller, Garny,
  Terkildsen, Lawson, Britten, Hunter, and Nielsen]{Yu2011.Physiome}
T.~Yu, C.~M. Lloyd, D.~P. Nickerson, M.~T. Cooling, A.~K. Miller, A.~Garny,
  J.~R. Terkildsen, J.~Lawson, R.~D. Britten, P.~J. Hunter, and P.~M.~F.
  Nielsen.
\newblock The {{Physiome Model Repository}} 2.
\newblock \emph{Bioinformatics}, 27\penalty0 (5):\penalty0 743--744, Mar. 2011.
\newblock ISSN 1367-4803.
\newblock \doi{10.1093/bioinformatics/btq723}.

\bibitem[Zeng et~al.(2023)Zeng, Chen, Zhang, and Xu]{Zeng2023.Are}
A.~Zeng, M.~Chen, L.~Zhang, and Q.~Xu.
\newblock Are {{Transformers Effective}} for {{Time Series Forecasting}}?
\newblock In \emph{{{AAAI}}}, 2023.

\bibitem[Zhang et~al.(2024)Zhang, Yin, Liu, Zhou, and
  Xiong]{Zhang2024.Irregular}
W.~Zhang, C.~Yin, H.~Liu, X.~Zhou, and H.~Xiong.
\newblock Irregular {{Multivariate Time Series Forecasting}}: {{A Transformable
  Patching Graph Neural Networks Approach}}.
\newblock In \emph{Proceedings of the 41st {{International Conference}} on
  {{Machine Learning}}}, pages 60179--60196. PMLR, July 2024.

\bibitem[Zhou et~al.(2021)Zhou, Zhang, Peng, Zhang, Li, Xiong, and
  Zhang]{Zhou2021.Informera}
H.~Zhou, S.~Zhang, J.~Peng, S.~Zhang, J.~Li, H.~Xiong, and W.~Zhang.
\newblock Informer: {{Beyond}} efficient transformer for long sequence
  time-series forecasting.
\newblock In \emph{Proceedings of the {{AAAI}} Conference on Artificial
  Intelligence}, volume~35, pages 11106--11115, 2021.

\end{thebibliography}

\clearpage\appendix
\clearpage\section{Real World Dataset Statistics}\label{sec:rw_dsinfo}

\begin{table}[H]
\centering
\caption{%
	Statistics of evaluation datasets used by~\citet{DeBrouwer2019.GRUODEBayesd,Bilos2021.Neurald,Schirmer2022.Modelingb}.
	\emph{Max. Len.} refers to the maximum sequence length among samples.
	\emph{Max. Obs.} refers to the maximum number of non-missing observations among samples.
	\emph{Sparsity} refers to the percentage of missing values over all samples.
}\label{tab:data}
\begin{tabular}{lccccc}
	\toprule
	name			& Instances & Channel & Max. Len & Max. Obs & Spars.
\\	\midrule
	USHCN			& 1.114    & 5        & 370      & 398      & 78.0\%
\\	PhysioNet-2012	& 11.981   & 37       & 48       & 606      & 80.4\%
\\	MIMIC-III		& 21.250   & 96       & 97       & 677      & 94.2\%
\\	MIMIC-IV		& 17.874   & 102      & 920      & 1642     & 97.8\%
\\	\bottomrule
\end{tabular}
\end{table}

\section{Proofs}\label{sec:proofs}
\subsection{Proof of Lemma~\ref{lemma:approx-tvl}}
\begin{proof}
Due to~\Cref{eq:mgd}, it is sufficient to show that
\begin{equation}\label{eq:convergence_numstd}
	\numstd[x^{\diff,\epsilon}]^2
	\xrightarrow{\epsilon \to 0}
	\frac{1}{T}\int_{t=0}^{T} \left(x'(t) - \frac{(x(T)-x(0))}{T}\right)^{2} \dd{t}.
\end{equation}
Due to the mean value theorem, there exists ${\zeta_{t,\epsilon} \in [t-\epsilon,t]}$
with ${\frac{x(t) - x(t-\epsilon)}{\epsilon} = x'(\zeta_{t,\epsilon})}$
for all ${t=\epsilon,2\epsilon,\dots,T}$.
Given ${N \coloneqq \frac{T}{\epsilon}}$ and ${\mu_N\coloneqq \frac{1}{N} \sum_{t=\epsilon,2\epsilon,\dots,T} x^{\diff,\epsilon}_{t}}$, we thus have
\begin{subequations}
\begin{align}
	\mu_N &= \frac{1}{N}\sum_{t=\epsilon,2\epsilon,\dots,T} x'(\zeta_{t,\epsilon})
\\	&=\frac{1}{T}\sum_{t=\epsilon,2\epsilon,\dots,T}\underset{=\epsilon}{\underbrace{\frac{T}{N}}} x'(\zeta_{t,\epsilon})
\\	&\xrightarrow{\epsilon \to 0} \frac{1}{T} \int_{t=0}^{T} x'(t) \dd{t}
\\	&= \frac{x(T)-x(0)}{T}
\end{align}
\end{subequations}
due to the definition of the Riemann integral. Furthermore, we have
\begin{align}
	\numstd[x^{\diff,\epsilon}]^2
	&= \frac{1}{N}\Bigl(\sum_{t=\epsilon,2\epsilon,\dots,T} (x^{\diff,\epsilon}_t)^{2}\Bigr)
	- \mu_N^{2}
\end{align}
Therefore, again by the mean value theorem and the definition of the Riemann integral:
\begin{subequations}
\begin{align}
	\frac{1}{N}\Bigl(\sum_{t=\epsilon,2\epsilon,\dots,T} (x^{\diff,\epsilon}_t)^{2}\Bigr)
	&= \frac{1}{N} \Bigl(\sum_{t=\epsilon,2\epsilon,\dots,T} x'(\zeta_{t,\epsilon})^{2}\Bigr)
\\	&= \frac{1}{T} \sum_{t=\epsilon,2\epsilon,\dots,T}  \underset{=\epsilon}{\underbrace{\frac{T}{N}}} x'(\zeta_{t,\epsilon})^{2}
\\	&\xrightarrow{\epsilon \to 0} \frac{1}{T} \int_{t=0}^{T} x'(t)^{2} \dd{t}
\end{align}
\end{subequations}
Consequently,
\begin{align}
	\numstd[x^{\diff,\epsilon}]^2
	&\xrightarrow{\epsilon \to 0}
	\frac{1}{T}\int_{t=0}^{T} x'(t)^{2} \dd{t} - \left(\frac{(x(T)-x(0))}{T}\right)^{2}
\\	&= \frac{1}{T}\int_{t=0}^{T} \left(x'(t) - \frac{(x(T)-x(0))}{T}\right)^{2} \dd{t}.
\end{align}
\end{proof}


\subsection{Assumptions and Proof of Lemma~\ref{lemma:approx-atv}}
For the lemma to hold we make the technical assumption:

(A) The distribution $p$ of continuous functions $[0,T]\to \R$ is given by 
a continuous map\footnote{%
	We define $\phi$ on the larger interval $[-1,T]$, instead of $[0,T]$, in order to use one-sided difference quotients.
} ${\phi\colon\Theta\times[-1,T]\to \R, (\theta,t)\mapsto \phi(\theta,t)}$,
with a compact parameter space $\Theta\subset \R^d$,
and a probability measure $\mu$ on $\Theta$.
We require that the map $\pdv{\phi}{t}$ is continuous, too.

We sometimes use $x_\theta$ for $\phi(\theta,\cdot\,)$  and $x'_\theta$ for $\pdv{\phi}{t}(\theta,\cdot\,)$.
This is no restriction for the differential
equations we consider, as we can always obtain a solution on $[-1,T]$, given parameters (including  $x(0)=x_0)$.

Sampling from this distribution is done by sampling from the probability measure on $\Theta$.
A sequence of i.i.d.\ random variables $\theta_n$, $n\in\N_{>0}$, distributed according to $\mu$, defines a sequence of random functions $(x_n)_{n\in\N_{>0}}$, $x_n(t) = \phi(\theta_n, t)$,
as in the statement of the \Cref{lemma:approx-atv}.

For the proof we will need a uniform law of large numbers (ULLN).
There exist many versions of such theorems
(e.g.\ \citep{jennrich_asymptotic_properties_non-linear_1969}, or \citep{andrews_consistency_1987}),
but we prefer to state and prove a version with strong assumptions appropriate for our setting.

\begin{lemma}[Uniform Law of Large Numbers]\label{lem:ulln}
Let $f\colon\Theta\times I \to \R$ be a continuous function on compact metric spaces $\Theta$ and $I$.
Let $\theta_i$, $i\in\N_{>0}$ be i.i.d.\ random variables with values in $\Theta$,
all distributed according to a probability measure $\mu$ on $\Theta$.
Then
\begin{align}
	\frac{1}{N}\sum_{i=1}^N  f(\theta_i, t)
	\xrightarrow{N\to\infty} \int_{\Theta} f(\theta, t) \dd{\mu(\theta)}
\end{align}
uniformly in $t\in I$ almost surely.
\end{lemma}

\begin{proof}\label{proof:ulln}
As $\Theta\times I$ is compact and $f$ continuous, $f$ is even uniformly continuous.
Hence, for any $\epsilon>0$, we can then choose a $\delta$ such that
\begin{align}
	d(\theta,\theta') < \delta
	\qq{and}
	(t,t')<\delta
	&\Longrightarrow \left|f(\theta',t')-f(\theta,t)\right|<\epsilon
\end{align}
Where $d$ is the metric on $\Theta$.
Now by compactness of $I$, finitely many $\delta$-balls in $I$ cover it,
i.e.\ there are ${t_1, \ldots, t_k\in I}$, such that the $B_\delta(t_j)$, ${j=1, \ldots, k}$ cover $I$.
For each $(t_j)_{j=1:k}$ we can apply a strong law of large numbers to obtain almost surely convergence of ${\frac{1}{N} \sum_{i=1}^N f(\theta_i, t_j)}$ to ${\int_{\Theta} f(\theta, t_j) \dd{\mu(\theta)}}$.

Hence, almost surely, the following holds: For $\epsilon>0$ we can choose $N_0>0$ such that
for all natural numbers $N\geq N_0$ and all $j\in\{1, \ldots, k\}$:
\begin{align}
	\left|
		\frac{1}{N} \sum_{i=1}^N f(\theta_i, t_j)
		- \int_{\Theta} f(\theta, t_j) \dd{\mu(\theta)}
	\right|
	&<\epsilon
\end{align}
Now for any $t\in I$ there is a $j\in\{1, \ldots, k\}$ such that
\begin{align}
	\left|\frac{1}{N} \sum_{i=1}^N \bigl(f(\theta_i, t_j) - f(\theta_i, t)\bigr)\right| &< \epsilon
	&&\text{and}&
	\left| \int_{\Theta} f(\theta, t_j) - f(\theta,t) \dd{\mu(\theta)}\right| &< \epsilon
\end{align}
The triangle inequality yields
\begin{align}
	\left|
		\frac{1}{N} \sum_{i=1}^N  f(\theta_i, t)
		-  \int_{\Theta} f(\theta, t) \dd{\mu(\theta)}
	\right|
	&< 3\epsilon
\end{align}
for all $t\in I$, which proves the claim.
\end{proof}

\begin{proof} (\Cref{lemma:approx-atv})
%
%
Let $\theta_i$, $i\in\N_{>0}$ be i.i.d.\ random variables
with values in $\Theta$, all distributed according to a probability measure $\mu$ on $\Theta$.

Define $\psi:\Theta\times[0,1]\times [0,T]$ by
\begin{align}
	\psi(\theta, \epsilon, t)
	&=
	\begin{cases}
		\frac{\phi(\theta, t)-\phi(\theta,t-\epsilon)}{\epsilon} & \text{if $\epsilon>0$}
	\\	\pdv{\phi}{t}(\theta, t) & \text{if $\epsilon=0$}
	\end{cases}
\end{align}
For $\epsilon>0$ the function is obviously continuous as a composition of continuous functions.
Let now be $\epsilon=0$ and $t\in[0,T]$, $\theta\in\Theta$. We claim that $\psi$ is continuous
at $(\theta,0,t)$. $\pdv{\phi}{t}$ is uniformly continuous on its compact domain of definition.

Let $\delta$ be greater than zero. We can choose a $\gamma>0$ such that for ${d(\theta,\theta')<\gamma}$ and ${|t-t'|<\gamma}$ we have ${\left|\pdv{\phi}{t}(\theta', t') - \pdv{\phi}{t}(\theta, t)\right|<\delta}$.

Now chose ${(\theta', \epsilon', t')}$ with ${d(\theta,\theta')<\gamma}$ and $d(t,t')<\gamma/2$ and $0\leq\epsilon'<\gamma/2$.
If $\epsilon'=0$, we directly obtain
\begin{align}
	\left|\psi(\theta',0, t') - \psi(\theta, 0, t)\right|
	&=  \left|\pdv{\phi}{t}(\theta',t') - \pdv{\phi}{t}(\theta,t)\right|
	< \delta
\end{align}
If $\epsilon'>0$, the mean value theorem implied that there is a $\tau\in(t'-\epsilon',t')$ such that:
\begin{align}
	\left|\psi(\theta',\epsilon', t')-\psi(\theta, 0, t)\right|
	&=  \left|\pdv{\phi}{t}(\theta',\tau) - \pdv{\phi}{t}(\theta,t)\right|
	< \delta
\end{align}
as ${d(\tau, t) \leq d(\tau,t') + d(t',t) \leq \epsilon'+\gamma/2 < \gamma/2+\gamma/2 = \gamma}$.
This establishes the continuity of $\psi$.

Therefore, $\psi$ is also uniformly continuous on its compact domain of definition.

Define $\Psi\colon[0,1]\times[0,T]\to \R$ by
\begin{align}
	\Psi(\epsilon, t) &= \int_{\Theta} \psi(\theta, \epsilon, t) \dd{\mu(\theta)}
\end{align}
We also define $\Psi^N\colon[0,1]\times[0,T]\to \R$
\begin{align}
	\Psi^N(\epsilon, t) &= \frac{1}{N} \sum_{i=1}^{N}\psi(\theta_i, \epsilon, t)
\end{align}
The uniform law of large numbers shows that almost surely  $\Psi^N$ converges uniformly
to $\Psi$ for $N\to\infty$.

Analogously we define $\Xi\colon[0,1]\to \R$ by
\begin{align}
	\Xi(\epsilon, t) &= \int_{\Theta} \psi(\theta, \epsilon, t)^2 \dd{\mu(\theta)}
\end{align}
We also define $\Psi^N\colon[0,1]\times[0,T]\to \R$
\begin{align}
	\Xi^N(\epsilon, t) &= \frac{1}{N} \sum_{i=1}^{N}\psi(\theta_i, \epsilon, t)^2
\end{align}
Again, the uniform law of large numbers shows that almost surely  $\Xi^N$ converges uniformly
to $\Xi$ for $N\to\infty$. Hence,
\begin{align}
	\numstd[(\psi(\theta_i,\epsilon,t))_{i=(1:N)}]
	&= \sqrt{\Xi^N(\epsilon,t)-\Psi^N(\epsilon, t)^2}
\end{align}
converges almost surely uniformly to $\sqrt{\Xi(\epsilon, t)-\Psi(\epsilon, t)^2}=\std_{\theta\sim \mu}[\Psi(\epsilon,t)]$ for $N\to\infty$.

On the other hand  by uniform continuity
\begin{align}
	\numstd[(\psi(\theta_i,\epsilon,t))_{i=(1:N)}]
	&\xrightarrow[\text{uniformly}]{\epsilon\to 0} \numstd[(\psi(\theta_i,0,t))_{i=(1:N)}]
\\	\sqrt{\Xi(\epsilon, t)-\Psi(\epsilon, t)^2}
	&\xrightarrow[\text{uniformly}]{\epsilon\to 0} \std_{\theta\sim \mu}[(\Psi(0,t))]
	= \std_{x\sim p} x'(t)
\end{align}

Now, given some $\delta>0$, we can choose an $\epsilon'$ and almost surely an $N_0$ such that
for all $0<\epsilon<\epsilon'$ and $N\geq N_0$:
\begin{align}
	\left|
		\numstd[(\psi(\theta_i,\epsilon,t))_{i=(1:N)}]
		- \numstd[(\psi(\theta_i,0,t))_{i=(1:N)}]
	\right|_{C^0([0,T])} &\leq \delta
	\\
	\left|
		\numstd[(\psi(\theta_i,0,t))_{i=(1:N)}]  - \std_{x\sim p} x'(t)
	\right|_{C^0([0,T])}
	&\leq \delta
\end{align}
and for $t,t'\in [0,T]$, $|t-t'|<\epsilon$:
\begin{align}
	\left|
		\numstd[(\psi(\theta_i,\epsilon,t))_{i=(1:N)}]
		- \numstd[(\psi(\theta_i,\epsilon,t'))_{i=(1:N)}]
	\right|
	&\leq \delta
\end{align}
Therefore,
\begin{align}
	\left|
		\frac{1}{T} \int_0^T \numstd[(\psi(\theta_i,\epsilon,t))_{i=(1:N)}] \dd{t}
		- \frac{1}{T} \int_0^T \std_{x\sim p} x'(t) \dd{t}
	\right|
	&< 2\delta
\end{align}
and for $T/\epsilon\in\N$ the Riemann sum with step size $\epsilon$ differs from the integral by at most $T\delta$, i.e.
\begin{align}
	\left|
		\frac{1}{T} \int_0^T \numstd[(\psi(\theta_i,\epsilon,t))_{i=(1:N)}] \dd{t}
		- \frac{\epsilon}{T} \sum_{k=1}^{T/\epsilon} \numstd[(\psi(\theta_i,\epsilon,k\epsilon))_{i=(1:N)}]
	\right|
	&<\delta
\end{align}
In the notations of the lemma, with $\frac{1}{T} \int_0^T \std_{x\sim p} x'(t) \dd{t} = \MPGD(p)$ we get:
\begin{align}
	\left|
		\frac{\epsilon}{T} \sum_{t=\epsilon}^{T} \numstd[(x^{\diff}_{i,t})_{i=1:N}] - \MPGD(p)
	\right|
	&< 3\delta
\end{align}
Thus proving the claim that the left term of the difference converges almost surely for $\epsilon\to 0 $ and $N\to\infty$ to the right term of the difference.
Please note that $N$ and $\epsilon$ were chosen independently. The limits can be taken simultaneously or in arbitrary order.
%
\end{proof}

\clearpage\section{Relation of MGD and Lipschitz Constant}\label{app:lip}

Regarding the relationship between the MGD and the Lipschitz constant of the vector field, we can consider the standard example of a linear ODE: $\dot{x}(t) = a \cdot x(t)$, for which the Lipschitz constant is precisely $a$. For simplicity, assume the initial condition $x(0) = 1$. Hence, the solution is $x(t) = e^{at}$. Then, by (4) we have $c = \frac{1}{T}(e^{aT} - 1)$, and hence:

\[
\begin{aligned}
\frac{1}{T} \int_0^T \|\dot{x}(t) - c\|^2 \, dt
&= \left( \frac{a}{2T} - \frac{1}{T^2} \right)e^{2aT} + \frac{1}{2T^2} e^{aT} - \frac{a}{2T} - \frac{1}{T^2}
\\&= \mathcal{O}\left(\frac{a}{T} e^{2aT}\right)
\\ \implies \text{MGD} &= \mathcal{O}\left(\sqrt{\frac{a}{T}} e^{aT}\right)
\end{aligned}
\]

In particular, for this example, the MGD grows super-exponentially with the Lipschitz constant. This result can be extended to an upper bound for general systems:

\begin{enumerate}
    \item By the Mean Value Theorem, there exists $\xi \in (0, T)$ such that $c = \frac{x(T) - x(0)}{T} = \dot{x}(\xi) = f(x(\xi))$.
    \item Therefore, $\|\dot{x}(t) - c\|^2 = \|\dot{x}(t) - \dot{x}(\xi)\|^2 = \|f(x(t)) - f(x(\xi))\|^2 \leq L^2 \|x(t) - x(\xi)\|^2$.
    \item We can bound the latter quantity via Grönwall's Lemma:
    \[
    \begin{aligned}
    \|x(t) - x(\xi)\|
    &= \left\|\int_{\xi}^t \dot{x}(s) \, ds \right\|
    \\&= \left\|\int_{\xi}^t f(x(s)) \, ds \right\|
    \\&\leq \int_{\xi}^t \|f(x(s))\| \, ds \quad \text{assuming } t \geq \xi
    \\&\leq \int_{\xi}^t \|f(x(s)) - f(x(\xi))\| + \|f(x(\xi))\| \, ds
    \\&\leq \underbrace{(t - \xi) \cdot \|f(x(\xi))\|}_{=\alpha(t)} + \int_{\xi}^t \underbrace{L}_{=\beta(s)} \cdot \underbrace{\|x(s) - x(\xi)\|}_{=u(s)} \, ds
    \\&\leq (t - \xi) \cdot \|f(x(\xi))\| \cdot e^{L(t - \xi)} \quad \text{via Grönwall}
    \end{aligned}
    \]

    And if $\xi \geq t$, we analogously get $\|x(t) - x(\xi)\| \leq (\xi - t) \cdot \|f(x(\xi))\| \cdot e^{L(\xi - t)}$.

    In particular, note that by the same method:
    \[
    \begin{aligned}
    \|f(x(\xi))\| = \frac{1}{T} \|x(T) - x(0)\| \leq \|f(x(0))\| e^{LT}
    \end{aligned}
    \]
\end{enumerate}

\begin{enumerate}
    \setcounter{enumi}{4}
    \item We use this result to bound the integral:
    \[
    \begin{aligned}
    \frac{1}{T} \int_0^T \|\dot{x}(t) - c\|^2 \, dt
    &\leq \frac{L^2}{T} \int_0^T \|x(t) - x(\xi)\|^2 \, dt
    \\&= \frac{L^2}{T} \left( \int_0^\xi \|x(t) - x(\xi)\|^2 \, dt + \int_\xi^T \|x(t) - x(\xi)\|^2 \, dt \right)
    \\&\leq \frac{L^2}{T} \|f(x(\xi))\|^2 \cdot \left( \int_0^\xi (\xi - t)^2 \cdot e^{2L(\xi - t)} \, dt + \int_\xi^T (t - \xi)^2 \cdot e^{2L(t - \xi)} \, dt \right)
    \\&\leq \dots
    \\&\leq \|f(x(\xi))\|^2 \left( (2LT^2 - T + \frac{1}{2L}) e^{2LT} - \frac{1}{2L} \right)
    \\&\lesssim \|f(x(0))\|^2 e^{2LT} \mathcal{O}(LT^2 e^{2LT})
    \\&= \mathcal{O}\left(\|f(x(0))\|^2 LTe^{4LT}\right)
    \end{aligned}
    \]
\end{enumerate}

\begin{enumerate}
    \setcounter{enumi}{5}
    \item Plugging this back into the definition of the MGD, we get:
    \[
    \begin{aligned}
    \text{MGD} &= \sqrt{\frac{1}{T} \int_0^T \|\dot{x}(t) - c\|^2 \, dt} = \mathcal{O}(\sqrt{LT} e^{2LT})
    \end{aligned}
    \]
\end{enumerate}

\clearpage\section{Dataset Description}\label{sec:dsinfo}
{\scriptsize\begin{longtable}{llrrlllll}
\caption{%
Optimized Datasets created with generated Python code published on Physiome. A t-unit of “---” refers to Physiome giving the time unit as dimensionless.
We entered --- in $\sigma_\text{dur}$, $\sigma_\text{state}$, $\sigma_\text{const}$ and $\JGD$, if we were unable to create a dataset with our method due to errors thrown by the ode solver or the occurence of ode-explosions (see~\Cref{sec:explosions})
}\label{tab:ds-info}
\\ \toprule
Model &                          Domain &  Channel &  Constants & Time Unit &    T & d-state & d-const &  ATVL \\
\midrule
\endfirsthead

\toprule
Model &                          Domain &  Channel &  Constants & Time Unit &    T & d-state & d-const &  ATVL \\
\midrule
\endhead
\midrule
\multicolumn{9}{r}{{Continued on next page}} \\
\midrule
\endfoot

\bottomrule
\endlastfoot
DUP01 &                Calcium-Dynamics &        2 &         13 &         s & 30.0 &     0.3 &    0.05 & 2.697 \\
JEL01 &                       Endocrine &        2 &         15 &         - & 30.0 &     0.5 &    0.05 &  2.58 \\
DOK01 &               Electrophysiology &       18 &         46 &         s & 10.0 &     0.1 &    0.05 & 2.277 \\
INA01 &               Electrophysiology &       29 &         58 &         s & 10.0 &     0.5 &    0.05 & 2.218 \\
WOL01 &             Signal-Transduction &        9 &         19 &       min & 10.0 &     0.3 &    0.05 & 1.973 \\
BOR01 &                Calcium-Dynamics &        3 &         14 &       min & 30.0 &     0.1 &    0.05 & 1.795 \\
HYN01 &                      Metabolism &       22 &         60 &       min & 30.0 &     0.1 &    0.05 & 1.548 \\
JEL02 &                       Endocrine &        2 &         15 &         - & 30.0 &     0.5 &    0.05 & 1.271 \\
DUP02 &                Calcium-Dynamics &        3 &         19 &       min & 30.0 &     0.1 &    0.05 & 1.202 \\
WOL02 &                      Metabolism &        7 &         14 &       min & 30.0 &     0.1 &     0.1 & 0.895 \\
DIF01 &               Electrophysiology &       16 &         50 &         s & 10.0 &     0.1 &     0.1 & 0.735 \\
VAN01 &                      Metabolism &       10 &         38 &         s & 10.0 &     0.1 &    0.05 & 0.407 \\
DUP03 &                Calcium-Dynamics &        2 &         14 &         s & 10.0 &     0.3 &    0.05 & 0.254 \\
BER01 &             Signal-Transduction &       11 &         22 &        ms & 30.0 &     0.1 &     0.3 & 0.179 \\
LEN01 &                       Endocrine &        3 &          7 &         s & 30.0 &     0.1 &     0.1 & 0.178 \\
 LI01 &               Electrophysiology &        5 &         20 &         s & 30.0 &     0.3 &    0.05 & 0.113 \\
 LI02 &               Electrophysiology &        5 &         20 &         s & 10.0 &     0.1 &    0.05 & 0.097 \\
REV01 &                      Immunology &        5 &         10 &       day & 30.0 &     0.5 &     0.3 & 0.081 \\
PUR01 &                    Neurobiology &        3 &         21 &        ms & 30.0 &     0.1 &    0.05 & 0.049 \\
NYG01 &               Electrophysiology &       29 &         51 &         s & 0.33 &     0.5 &     0.1 & 0.047 \\
PUR02 &                    Neurobiology &        3 &         21 &        ms & 30.0 &     0.1 &    0.05 & 0.044 \\
HOD01 &               Electrophysiology &        4 &          8 &        ms & 30.0 &     0.5 &     0.3 & 0.042 \\
REE01 &                      Metabolism &        4 &         21 &         h & 30.0 &     0.5 &    0.05 & 0.035 \\
VIL01 &             Signal-Transduction &        9 &         16 &         h & 30.0 &     0.1 &     0.3 & 0.028 \\
KAR01 &             Signal-Transduction &       16 &         21 &         s & 10.0 &     0.3 &    0.05 & 0.023 \\
SHO01 & Excitation-contraction-Coupling &       56 &        105 &        ms & 30.0 &     0.3 &    0.05 & 0.023 \\
BUT01 &               Electrophysiology &        4 &         24 &        ms & 30.0 &     0.3 &    0.05 &  0.02 \\
MAL01 &                       Endocrine &        9 &         51 &       day & 30.0 &     0.5 &    0.05 & 0.019 \\
ASL01 &               Electrophysiology &       30 &         83 &        ms & 30.0 &     0.3 &     0.3 & 0.019 \\
BUT02 &               Electrophysiology &        4 &         25 &        ms & 30.0 &     0.3 &    0.05 & 0.016 \\
MIT01 &                      Immunology &        7 &         13 &       day & 30.0 &     0.1 &     0.3 & 0.015 \\
GUP01 &                       Endocrine &        4 &          7 &         h & 30.0 &     0.5 &    0.05 & 0.014 \\
GUY01 &      Cardiovascular-Circulation &        2 &          9 &       min & 30.0 &     0.3 &    0.05 & 0.013 \\
PHI01 &               Circadian-Rhythms &        3 &         17 &         h & 30.0 &     0.1 &     0.1 & 0.013 \\
GUY02 &      Cardiovascular-Circulation &        2 &         11 &       min & 30.0 &     0.3 &    0.05 & 0.013 \\
PUL01 &      Cardiovascular-Circulation &        2 &        561 &       min & 30.0 &     0.3 &    0.05 & 0.013 \\
CAL01 &                      Cell-Cycle &       17 &         43 &       min & 30.0 &     0.5 &     0.1 & 0.013 \\
WOD01 &                      Immunology &        3 &          7 &         s & 30.0 &     0.1 &     0.1 & 0.012 \\
GUP02 &                       Endocrine &        4 &          7 &         h & 30.0 &     0.5 &     0.1 & 0.012 \\
  M01 &      Cardiovascular-Circulation &        2 &        558 &       min & 30.0 &     0.3 &    0.05 & 0.012 \\
LEN02 &                       Endocrine &        3 &         12 &         s & 30.0 &     0.5 &     0.1 & 0.012 \\
KAR02 &             Signal-Transduction &       13 &         15 &         s &  3.3 &     0.1 &     0.1 & 0.011 \\
SHO02 & Excitation-contraction-Coupling &       56 &        105 &        ms & 10.0 &     0.3 &    0.05 & 0.011 \\
MAC01 &                   Ion-Transport &        7 &         35 &         s & 0.33 &     0.5 &    0.05 &  0.01 \\
IRI01 &               Electrophysiology &       23 &         69 &         s & 0.33 &     0.3 &     0.1 &  0.01 \\
BAG01 &             Signal-Transduction &        9 &         30 &         s &  3.3 &     0.1 &    0.05 &  0.01 \\
WOL03 &                      Metabolism &       13 &         28 &       min & 30.0 &     0.5 &     0.1 & 0.008 \\
WAN01 &                Calcium-Dynamics &        5 &         27 &         s & 30.0 &     0.3 &     0.1 & 0.008 \\
NEL01 &                      Immunology &        4 &          7 &       day & 30.0 &     0.1 &     0.3 & 0.007 \\
HUA01 &             Signal-Transduction &       18 &         37 &         s & 30.0 &     0.1 &     0.3 & 0.007 \\
\hline
SVE01 &                      Cell-Cycle &       16 &         58 &       min & 10.0 &     0.5 &     0.3 & 0.007 \\
GUY03 &      Cardiovascular-Circulation &        2 &          8 &       min & 30.0 &     0.5 &    0.05 & 0.006 \\
CIL01 &                      Cell-Cycle &       19 &         69 &       min & 10.0 &     0.3 &    0.05 & 0.006 \\
LEL01 &             Signal-Transduction &       10 &         38 &         h & 30.0 &     0.5 &    0.05 & 0.006 \\
GAR01 &                      Cell-Cycle &        5 &         20 &       min & 30.0 &     0.5 &     0.3 & 0.006 \\
COU01 &               Electrophysiology &       21 &         49 &        ms & 30.0 &     0.5 &    0.05 & 0.005 \\
AUT01 &      Cardiovascular-Circulation &        2 &        570 &       min &  3.3 &     0.5 &    0.05 & 0.005 \\
GUY04 &      Cardiovascular-Circulation &        2 &         41 &       min &  3.3 &     0.5 &    0.05 & 0.005 \\
RAZ01 &           Myofilament-Mechanics &        3 &         19 &         s & 30.0 &     0.1 &     0.1 & 0.005 \\
GRA01 &                            Pkpd &        9 &         30 &         h & 10.0 &     0.5 &    0.05 & 0.005 \\
SNE01 &                Calcium-Dynamics &        5 &         26 &         s &  1.0 &     0.5 &     0.3 & 0.005 \\
WOD02 &                      Immunology &       10 &         14 &       day & 10.0 &     0.3 &     0.1 & 0.005 \\
WOD03 &                      Immunology &        4 &         12 &       day & 30.0 &     0.5 &     0.1 & 0.004 \\
NEU01 &                      Immunology &        3 &          8 &       day & 10.0 &     0.3 &     0.3 & 0.004 \\
FAR01 &             Signal-Transduction &        5 &         35 &         h &  1.0 &     0.1 &     0.1 & 0.004 \\
SNE02 &                Calcium-Dynamics &        3 &         13 &         s & 30.0 &     0.3 &     0.1 & 0.004 \\
FAR02 &             Signal-Transduction &        5 &         35 &         h &  1.0 &     0.1 &     0.1 & 0.004 \\
BER02 &               Electrophysiology &        7 &         47 &        ms & 30.0 &     0.5 &     0.1 & 0.004 \\
GUY05 &      Cardiovascular-Circulation &        3 &         13 &       min & 30.0 &     0.3 &    0.05 & 0.004 \\
MCA01 &               Electrophysiology &       10 &         13 &        ms & 30.0 &     0.3 &     0.3 & 0.004 \\
MCA02 &               Electrophysiology &       10 &         13 &        ms & 30.0 &     0.3 &     0.3 & 0.004 \\
NOV01 &                      Cell-Cycle &       11 &         45 &       min & 30.0 &     0.5 &     0.3 & 0.004 \\
YAM01 & Excitation-contraction-Coupling &        5 &         20 &         s &  3.3 &     0.1 &     0.1 & 0.004 \\
RAL01 &                      Metabolism &       24 &        142 &       min & 30.0 &     0.1 &     0.3 & 0.004 \\
VIS01 &               Electrophysiology &       25 &         75 &        ms & 10.0 &     0.5 &    0.05 & 0.004 \\
CAM01 &           Myofilament-Mechanics &        4 &         10 &         s & 30.0 &     0.3 &    0.05 & 0.004 \\
VIS02 &               Electrophysiology &       25 &         75 &        ms & 10.0 &     0.5 &    0.05 & 0.004 \\
SNY01 &                Calcium-Dynamics &       10 &         27 &         s & 30.0 &     0.5 &     0.3 & 0.003 \\
NOB01 &               Electrophysiology &       20 &         69 &         s & 0.33 &     0.1 &    0.05 & 0.003 \\
LEL02 &             Signal-Transduction &        3 &         10 &         h & 30.0 &     0.5 &     0.3 & 0.003 \\
MAL02 &               Electrophysiology &       29 &         86 &        ms & 30.0 &     0.5 &    0.05 & 0.003 \\
UED01 &             Signal-Transduction &       10 &         55 &         h & 30.0 &     0.5 &     0.1 & 0.002 \\
CHE01 &                      Cell-Cycle &       36 &        136 &       min & 10.0 &     0.5 &     0.3 & 0.002 \\
POT01 &                       Endocrine &       33 &         97 &         h & 30.0 &     0.1 &     0.1 & 0.002 \\
CHE02 &                      Cell-Cycle &       13 &         71 &       min & 30.0 &     0.1 &    0.05 & 0.002 \\
PAR01 &                   Ion-Transport &        7 &         35 &         s &  3.3 &     0.5 &    0.05 & 0.002 \\
JEL03 &                       Endocrine &        2 &         15 &         - & 30.0 &     0.1 &     0.1 & 0.002 \\
GOO01 &               Circadian-Rhythms &        2 &          6 &         s & 30.0 &     0.5 &     0.3 & 0.002 \\
DIX01 &                      Immunology &        5 &         18 &       day &  3.3 &     0.5 &    0.05 & 0.002 \\
CSI01 &                      Cell-Cycle &       14 &         93 &       min & 10.0 &     0.5 &     0.1 & 0.002 \\
HAT01 &             Signal-Transduction &       33 &         80 &         s & 30.0 &     0.5 &     0.1 & 0.002 \\
VAS01 &                      Metabolism &       10 &         19 &         s & 30.0 &     0.1 &     0.3 & 0.002 \\
MIT02 &               Electrophysiology &        2 &         10 &        ms & 30.0 &     0.3 &     0.3 & 0.002 \\
MIJ01 &           Myofilament-Mechanics &        4 &         12 &         s & 30.0 &     0.5 &     0.3 & 0.002 \\
REI01 &             Signal-Transduction &        4 &         11 &         s &  1.0 &     0.1 &     0.3 & 0.002 \\
GUY06 &      Cardiovascular-Circulation &        5 &         45 &       min &  3.3 &     0.3 &    0.05 & 0.002 \\
GAL01 &               Electrophysiology &        3 &         28 &        ms & 30.0 &     0.5 &    0.05 & 0.002 \\
LEM01 &                       Endocrine &        3 &         28 &       day & 30.0 &     0.5 &    0.05 & 0.002 \\
SRI01 &                      Cell-Cycle &        6 &         25 &       min & 30.0 &     0.5 &    0.05 & 0.001 \\
FAB01 &               Electrophysiology &       25 &         81 &        ms & 10.0 &     0.3 &    0.05 & 0.001 \\
CUI01 &                Calcium-Dynamics &        4 &         22 &       min &  3.3 &     0.1 &     0.3 & 0.001 \\
LEM02 &                       Endocrine &        3 &         28 &       day & 30.0 &     0.5 &    0.05 & 0.001 \\
SCH01 &                      Immunology &        5 &         15 &       day & 30.0 &     0.5 &     0.3 & 0.001 \\
HAU01 &                       Endocrine &        4 &         14 &       min & 30.0 &     0.5 &     0.1 & 0.001 \\
BOR02 &                Calcium-Dynamics &        3 &         15 &       min & 0.33 &     0.3 &    0.05 & 0.001 \\
WOD04 &                      Immunology &        2 &          5 &         s & 30.0 &     0.5 &     0.3 & 0.001 \\
FIN01 &               Electrophysiology &       27 &         61 &        ms & 0.33 &     0.1 &     0.3 & 0.001 \\
CHA01 &                      Metabolism &       12 &         43 &       min & 0.33 &     0.1 &     0.3 & 0.001 \\
HAL01 &                      Cell-Cycle &        4 &          7 &     month & 30.0 &     0.1 &    0.05 & 0.001 \\
BAR01 &                      Cell-Cycle &       34 &         55 &       min &  3.3 &     0.1 &    0.05 & 0.001 \\
RAP01 &                Calcium-Dynamics &       11 &         57 &         h & 10.0 &     0.5 &    0.05 & 0.001 \\
KOM01 &                      Immunology &        2 &          9 &         h & 30.0 &     0.5 &     0.1 & 0.001 \\
NOV02 &                      Cell-Cycle &        9 &         31 &       min & 10.0 &     0.5 &    0.05 & 0.001 \\
NOV03 &                      Cell-Cycle &       13 &         43 &       min & 10.0 &     0.5 &    0.05 & 0.001 \\
VEM01 &             Signal-Transduction &       17 &         33 &         s & 30.0 &     0.5 &     0.3 & 0.001 \\
MIF01 &               Electrophysiology &        5 &         14 &        ms & 0.33 &     0.1 &     0.3 & 0.001 \\
COR01 &               Electrophysiology &       14 &         51 &         - & 30.0 &     0.5 &    0.05 & 0.001 \\
BON01 &                      Immunology &        3 &          7 &       day & 30.0 &     0.5 &     0.3 & 0.001 \\
YAT01 &                      Cell-Cycle &        2 &          5 &       day & 30.0 &     0.5 &    0.05 & 0.001 \\
BON02 &                      Immunology &        4 &         13 &       day & 30.0 &     0.5 &     0.3 & 0.001 \\
BON03 &                      Immunology &        3 &          8 &       day & 30.0 &     0.5 &     0.3 & 0.001 \\
BRO01 &                       Endocrine &        2 &         11 &       min & 30.0 &     0.5 &     0.3 &   0.0 \\
IZA01 & Excitation-contraction-Coupling &        2 &         30 &        ms & 10.0 &     0.1 &     0.3 &   0.0 \\
VIL02 &             Signal-Transduction &        6 &          9 &       min & 30.0 &     0.5 &    0.05 &   0.0 \\
BER03 &               Electrophysiology &        4 &         21 &        ms & 30.0 &     0.1 &     0.3 &   0.0 \\
BER04 &               Electrophysiology &        4 &         21 &        ms & 30.0 &     0.1 &     0.3 &   0.0 \\
BER05 &               Electrophysiology &        5 &         32 &        ms & 30.0 &     0.5 &    0.05 &   0.0 \\
KOM02 &                       Endocrine &        3 &         13 &       day & 30.0 &     0.1 &     0.1 &   0.0 \\
LEN03 &                       Endocrine &        5 &         15 &         s & 30.0 &     0.5 &     0.3 &   0.0 \\
RAT01 &                       Endocrine &        3 &         13 &         - & 30.0 &     0.5 &     0.1 &   0.0 \\
BUE01 &               Electrophysiology &        4 &         32 &        ms & 30.0 &     0.5 &     0.1 &   0.0 \\
GUY07 &      Cardiovascular-Circulation &        4 &         27 &       min & 30.0 &     0.5 &    0.05 &   0.0 \\
HUN01 &               Electrophysiology &       29 &         70 &        ms &  1.0 &     0.1 &     0.1 &   0.0 \\
ROM01 &                      Cell-Cycle &        6 &         28 &       min & 30.0 &     0.5 &     0.1 &   0.0 \\
SHO03 &                Calcium-Dynamics &        7 &         40 &        ms &  1.0 &     0.1 &     0.1 &   0.0 \\
GUY08 &      Cardiovascular-Circulation &        2 &         17 &       min & 30.0 &     0.5 &    0.05 &   0.0 \\
POT02 &                       Endocrine &        5 &         12 &         s & 30.0 &     0.5 &    0.05 &   0.0 \\
CIR01 &      Cardiovascular-Circulation &       44 &         96 &        ms & 30.0 &     0.1 &    0.05 &   0.0 \\
STE01 & Excitation-contraction-Coupling &        4 &          5 &        ms & 30.0 &     0.5 &     0.3 &   0.0 \\
PED01 &                       Endocrine &        9 &         54 &        ms & 10.0 &     0.3 &    0.05 &   0.0 \\
GUY09 &      Cardiovascular-Circulation &        2 &          5 &       min & 30.0 &     0.1 &     0.3 &   0.0 \\
GON01 &                       Endocrine &        3 &         11 &       min & 0.33 &     0.1 &    0.05 &   0.0 \\
ELE01 &      Cardiovascular-Circulation &        4 &        566 &       min & 30.0 &     0.1 &    0.05 &   0.0 \\
GUY10 &      Cardiovascular-Circulation &        4 &         17 &       min & 30.0 &     0.1 &     0.3 &   0.0 \\
POT03 &                       Endocrine &        5 &         16 &         s & 30.0 &     0.1 &     0.3 &   0.0 \\
CLO01 &                      Metabolism &        7 &         42 &         s & 0.33 &     0.1 &     0.1 &   0.0 \\
CLO02 &                      Metabolism &        7 &         42 &         s & 0.33 &     0.1 &     0.1 &   0.0 \\
RAN01 &             Signal-Transduction &       32 &         32 &         s & 30.0 &     0.1 &     0.3 &   0.0 \\
HYP01 &      Cardiovascular-Circulation &        3 &        560 &       min & 30.0 &     0.1 &     0.3 &   0.0 \\
GUY11 &      Cardiovascular-Circulation &        3 &         10 &       min & 30.0 &     0.1 &     0.3 &   0.0 \\
SED01 &             Signal-Transduction &       20 &         33 &       min &    - &       - &       - &     - \\
LIV01 &               Electrophysiology &       18 &         68 &        ms &    - &       - &       - &     - \\
KAT01 &               Electrophysiology &        9 &         54 &         s &    - &       - &       - &     - \\
CUI02 &               Electrophysiology &       20 &         17 &         s &    - &       - &       - &     - \\
NIE01 &               Electrophysiology &       26 &        188 &        ms &    - &       - &       - &     - \\
CHA02 &               Electrophysiology &       12 &         38 &         s &    - &       - &       - &     - \\
KAT02 &               Electrophysiology &        9 &         54 &        ms &    - &       - &       - &     - \\
FAV01 &               Electrophysiology &       80 &        770 &         s &    - &       - &       - &     - \\
CHA03 &               Electrophysiology &        9 &         27 &         s &    - &       - &       - &     - \\
CLO03 &                    Neurobiology &       36 &        125 &         s &    - &       - &       - &     - \\
VIN01 &                      Metabolism &       25 &        342 &       min &    - &       - &       - &     - \\
DEM01 &             Signal-Transduction &       29 &         55 &         s &    - &       - &       - &     - \\
CHE03 &             Signal-Transduction &       14 &         20 &         s &    - &       - &       - &     - \\
COO01 &             Signal-Transduction &       13 &         38 &         s &    - &       - &       - &     - \\
NAZ01 &                      Metabolism &        8 &         55 &         - &    - &       - &       - &     - \\
HUN02 &               Electrophysiology &       29 &         69 &        ms &    - &       - &       - &     - \\
NAZ02 &                      Metabolism &        8 &         25 &         s &    - &       - &       - &     - \\
WAN02 &             Signal-Transduction &        4 &          4 &         s &    - &       - &       - &     - \\
CUR01 &                      Metabolism &        0 &         21 &           &    - &       - &       - &     - \\
DAS01 &                      Metabolism &        0 &         30 &           &    - &       - &       - &     - \\
VIN02 &                      Metabolism &       25 &        342 &       min &    - &       - &       - &     - \\
HEY01 &                      Metabolism &       11 &         18 &         s &    - &       - &       - &     - \\
CHE04 &             Signal-Transduction &       14 &         20 &         s &    - &       - &       - &     - \\
PRO01 &             Signal-Transduction &       14 &         24 &         s &    - &       - &       - &     - \\
TRA01 &                      Metabolism &       11 &         71 &        ms &    - &       - &       - &     - \\
BEA01 &                      Metabolism &       19 &         60 &         s &    - &       - &       - &     - \\
BEN01 &               Electrophysiology &       29 &         69 &        ms &    - &       - &       - &     - \\
NOB02 &               Electrophysiology &       22 &         70 &         s &    - &       - &       - &     - \\
 LI03 &               Electrophysiology &       12 &         37 &         s &    - &       - &       - &     - \\
THA01 &                            Pkpd &        2 &         15 &      week &    - &       - &       - &     - \\
MAC02 &                       Endocrine &        4 &         14 &       min &    - &       - &       - &     - \\
MAI01 &      Cardiovascular-Circulation &        7 &         58 &         s &    - &       - &       - &     - \\
GUY12 &      Cardiovascular-Circulation &       43 &        210 &       min &    - &       - &       - &     - \\
MAI02 &      Cardiovascular-Circulation &        3 &         21 &         s &    - &       - &       - &     - \\
KYR01 &                       Endocrine &        5 &         24 &       min &    - &       - &       - &     - \\
MAC03 &                       Endocrine &        7 &         28 &       min &    - &       - &       - &     - \\
BAY01 &                Calcium-Dynamics &        3 &          2 &         s &    - &       - &       - &     - \\
RIC01 &             Signal-Transduction &        8 &         11 &         s &    - &       - &       - &     - \\
BEN02 &    Mechanical-Constitutive-Laws &        7 &         26 &         s &    - &       - &       - &     - \\
PAN01 & Excitation-contraction-Coupling &       22 &         93 &        ms &    - &       - &       - &     - \\
MAI03 &      Cardiovascular-Circulation &        7 &         32 &         s &    - &       - &       - &     - \\
OVE01 &                            Pkpd &        3 &         37 &         h &    - &       - &       - &     - \\
PMR01 &                 Gene-Regulation &        0 &          0 &           &    - &       - &       - &     - \\
GUY13 &      Cardiovascular-Circulation &        0 &         10 &           &    - &       - &       - &     - \\
MOD01 &      Cardiovascular-Circulation &       14 &         88 &         s &    - &       - &       - &     - \\
PMR02 &                 Gene-Regulation &        0 &          0 &           &    - &       - &       - &     - \\
GUY14 &      Cardiovascular-Circulation &        4 &         72 &       min &    - &       - &       - &     - \\
ESP01 &               Electrophysiology &       21 &         70 &         s &    - &       - &       - &     - \\
BEN03 &               Electrophysiology &       29 &         69 &        ms &    - &       - &       - &     - \\
BOY01 &               Electrophysiology &       22 &         90 &         s &    - &       - &       - &     - \\
ESP02 &               Electrophysiology &       21 &         70 &         s &    - &       - &       - &     - \\
ASL02 &               Electrophysiology &       29 &         49 &         s &    - &       - &       - &     - \\
OST01 &               Electrophysiology &        8 &         31 &         s &    - &       - &       - &     - \\
MIC01 &               Electrophysiology &       43 &         97 &         s &    - &       - &       - &     - \\
NIE02 &               Electrophysiology &        5 &         25 &        ms &    - &       - &       - &     - \\
INA02 &               Electrophysiology &       29 &         63 &         s &    - &       - &       - &     - \\
MAH01 &               Electrophysiology &       26 &         80 &        ms &    - &       - &       - &     - \\
SED02 &             Signal-Transduction &       21 &         38 &       min &    - &       - &       - &     - \\
\end{longtable}}

\clearpage\section{Links of ODE models}\label{sec:ds_links}
{\scriptsize\begin{longtable}{ll}
\caption{%
Model abbreviations with links to their Physiome page. The Physiome page contains information about the ODE model, the respective publication. 
Furthermore, the generated Python code can also be found there.
} \\
\toprule
Abbreviation &                                                                               Link \\
\midrule
\endfirsthead

\toprule
Abbreviation &                                                                               Link \\
\midrule
\endhead
\midrule
\multicolumn{2}{r}{{Continued on next page}} \\
\midrule
\endfoot

\bottomrule
\endlastfoot
ASL01 & \url{https://models.physiomeproject.org/exposure/bb75b3021730966e2827967a66188cb2} \\
ASL02 & \url{https://models.physiomeproject.org/exposure/0e3a603db8f464ae89becb8d89225d90} \\
AUT01 & \url{https://models.physiomeproject.org/exposure/827af05888f8e152f448d9cd8c6a8d09} \\
BAG01 & \url{https://models.physiomeproject.org/exposure/98c3ef8d8c8f444d3a492fe777b4f6d0} \\
BAR01 & \url{https://models.physiomeproject.org/exposure/455155cfef38300290a75bcfe7440c31} \\
BAY01 & \url{https://models.physiomeproject.org/exposure/210f6601f6461be8443592ff071d2592} \\
BEA01 & \url{https://models.physiomeproject.org/exposure/959a4da5d8a0638f4b36adf5800f4fc0} \\
BEN01 & \url{https://models.physiomeproject.org/exposure/5d2fd39aa16f562d3bc19119847b7db0} \\
BEN02 & \url{https://models.physiomeproject.org/exposure/b503501533abcf0e70786789f08cb902} \\
BEN03 & \url{https://models.physiomeproject.org/exposure/5d2fd39aa16f562d3bc19119847b7db0} \\
BER01 & \url{https://models.physiomeproject.org/exposure/c6d3b19b455fbb3f80813337ac127674} \\
BER02 & \url{https://models.physiomeproject.org/exposure/a7f8b4574fe4802f1e06f247006962ac} \\
BER03 & \url{https://models.physiomeproject.org/exposure/fe91f06c8311ffd90fd1eaabe9b0032a} \\
BER04 & \url{https://models.physiomeproject.org/exposure/fe91f06c8311ffd90fd1eaabe9b0032a} \\
BER05 & \url{https://models.physiomeproject.org/exposure/48e767ee37f347e1a3876d0549a0866b} \\
BON01 & \url{https://models.physiomeproject.org/exposure/def2cd06391e17d8966e22997a65476a} \\
BON02 & \url{https://models.physiomeproject.org/exposure/def2cd06391e17d8966e22997a65476a} \\
BON03 & \url{https://models.physiomeproject.org/exposure/def2cd06391e17d8966e22997a65476a} \\
BOR01 & \url{https://models.physiomeproject.org/exposure/7ba127c00a5524c49f0a57e6589eda03} \\
BOR02 & \url{https://models.physiomeproject.org/exposure/7ba127c00a5524c49f0a57e6589eda03} \\
BOY01 & \url{https://models.physiomeproject.org/exposure/d4d2febdd2a9556eca88ceeac66bf696} \\
BRO01 & \url{https://models.physiomeproject.org/exposure/1f40088c3eb67895ff8dcc5010d62bf1} \\
BUE01 & \url{https://models.physiomeproject.org/exposure/954421556db2a0c7de89ce0bdc26bed9} \\
BUT01 & \url{https://models.physiomeproject.org/exposure/293a909eeeca07d0ca8ad583839eb0bc} \\
BUT02 & \url{https://models.physiomeproject.org/exposure/293a909eeeca07d0ca8ad583839eb0bc} \\
CAL01 & \url{https://models.physiomeproject.org/exposure/1a3f36d015121d5596565fe7d9afb332} \\
CAM01 & \url{https://models.physiomeproject.org/exposure/62183706711e435ff002b46088540850} \\
CHA01 & \url{https://models.physiomeproject.org/exposure/a74fd808df7ba454945e5ac42486e3ad} \\
CHA02 & \url{https://models.physiomeproject.org/exposure/1f4c6ffdb5120ed09675e4c9d346952f} \\
CHA03 & \url{https://models.physiomeproject.org/exposure/7fce899159b8698ccb2c8569d19a0ea1} \\
CHE01 & \url{https://models.physiomeproject.org/exposure/0e2337ffa4c550dd077c629b9e76330e} \\
CHE02 & \url{https://models.physiomeproject.org/exposure/635c1359f216ab8d064dfe8e8d3387b8} \\
CHE03 & \url{https://models.physiomeproject.org/exposure/01f8729559c3985a3d09ff4d983da75a} \\
CHE04 & \url{https://models.physiomeproject.org/exposure/01f8729559c3985a3d09ff4d983da75a} \\
CIL01 & \url{https://models.physiomeproject.org/exposure/f5be28698749563c5564d1a4fa5c0446} \\
CIR01 & \url{https://models.physiomeproject.org/exposure/ff8be5f140e68612284488cf9879eb5f} \\
CLO01 & \url{https://models.physiomeproject.org/exposure/e5cfb42225d4534a1e08979e57cf8bdd} \\
CLO02 & \url{https://models.physiomeproject.org/exposure/e5cfb42225d4534a1e08979e57cf8bdd} \\
CLO03 & \url{https://models.physiomeproject.org/exposure/6e249a04f5c751e42ba41504d84e6e49} \\
COO01 & \url{https://models.physiomeproject.org/exposure/39fee05138f428413db61aa29ff6f0de} \\
COR01 & \url{https://models.physiomeproject.org/exposure/9477b42c85e81f7da21059baa509ae21} \\
COU01 & \url{https://models.physiomeproject.org/exposure/0e03bbe01606be5811691f9d5de10b65} \\
CSI01 & \url{https://models.physiomeproject.org/exposure/dada9c9d39d0783375d42f12f11dddf3} \\
CUI01 & \url{https://models.physiomeproject.org/exposure/a3fb8998a3dd86ca914e55d60b3b7a52} \\
CUI02 & \url{https://models.physiomeproject.org/exposure/45ddea4fcde109d7a2bef806aeae84d7} \\
CUR01 & \url{https://models.physiomeproject.org/exposure/e08bd642df06eb59404c8e46afbb04a4} \\
DAS01 & \url{https://models.physiomeproject.org/exposure/bb0ee0e6d141a529fa896ff3785f0eff} \\
DEM01 & \url{https://models.physiomeproject.org/exposure/32c9e9739454b40b5ba2d9cabb1fd079} \\
DIF01 & \url{https://models.physiomeproject.org/exposure/91d93b61d7da56b6baf1f0c4d88ecd77} \\
DIX01 & \url{https://models.physiomeproject.org/exposure/cb02cc9ce006eb44b2077111c1d548d2} \\
DOK01 & \url{https://models.physiomeproject.org/exposure/462ab10275dfc099166c8a0e4f9e1be3} \\
DUP01 & \url{https://models.physiomeproject.org/exposure/060c119cc5365e3d9cd0203c82fe0121} \\
DUP02 & \url{https://models.physiomeproject.org/exposure/04cc3dc1c0c8e5dc3634f0220904d297} \\
DUP03 & \url{https://models.physiomeproject.org/exposure/060c119cc5365e3d9cd0203c82fe0121} \\
ELE01 & \url{https://models.physiomeproject.org/exposure/239e0bdec0c7054e0727fc111a54658c} \\
ESP01 & \url{https://models.physiomeproject.org/exposure/3ead51a9899108b4f8c8177c0b5f2421} \\
ESP02 & \url{https://models.physiomeproject.org/exposure/3ead51a9899108b4f8c8177c0b5f2421} \\
FAB01 & \url{https://models.physiomeproject.org/exposure/55643f2114a2a463ada007deb9fc3913} \\
FAR01 & \url{https://models.physiomeproject.org/exposure/fe9ed942c523777274fcdb43c6be75f7} \\
FAR02 & \url{https://models.physiomeproject.org/exposure/fe9ed942c523777274fcdb43c6be75f7} \\
FAV01 & \url{https://models.physiomeproject.org/exposure/f17a3cade0f0f7986b060b8bbf5c2957} \\
FIN01 & \url{https://models.physiomeproject.org/exposure/eeb81adc372c2f172399ec7160b0331e} \\
GAL01 & \url{https://models.physiomeproject.org/exposure/bc0d80f7e1ccb2c288119f6e8fdfbc20} \\
GAR01 & \url{https://models.physiomeproject.org/exposure/4f5fb67ce413238e24e5277f46fd5d21} \\
GON01 & \url{https://models.physiomeproject.org/exposure/07464e63da82057cc6e44d33ec60cd7c} \\
GOO01 & \url{https://models.physiomeproject.org/exposure/b69b4d821f5b2996cda5f2b4a3e9ca8f} \\
GRA01 & \url{https://models.physiomeproject.org/exposure/32d3dcc3ab074d17905c10f2ba26b54f} \\
GUP01 & \url{https://models.physiomeproject.org/exposure/9b91dc35df3c4ca3ee14f6a3f4191db7} \\
GUP02 & \url{https://models.physiomeproject.org/exposure/9b91dc35df3c4ca3ee14f6a3f4191db7} \\
GUY01 & \url{https://models.physiomeproject.org/exposure/d20c1a1db1b4027605b87b4361e62560} \\
GUY02 & \url{https://models.physiomeproject.org/exposure/63f124c018745aa8ec92a091434f3dfc} \\
GUY03 & \url{https://models.physiomeproject.org/exposure/c1461e469a5e282face2fdf37817ca10} \\
GUY04 & \url{https://models.physiomeproject.org/exposure/827af05888f8e152f448d9cd8c6a8d09} \\
GUY05 & \url{https://models.physiomeproject.org/exposure/be69ef6c1a42264162ae6e370e6107b5} \\
GUY06 & \url{https://models.physiomeproject.org/exposure/e1f152d0dbd1dec4111faa117db85bc0} \\
GUY07 & \url{https://models.physiomeproject.org/exposure/f3272a51c6e95c70eb1309d30c08d4cf} \\
GUY08 & \url{https://models.physiomeproject.org/exposure/f97a5eb092b12f4f0f32ac51ee20d20e} \\
GUY09 & \url{https://models.physiomeproject.org/exposure/64f53aa4552f15dc6805444ba0559c5a} \\
GUY10 & \url{https://models.physiomeproject.org/exposure/239e0bdec0c7054e0727fc111a54658c} \\
GUY11 & \url{https://models.physiomeproject.org/exposure/ea5144be46b69ba503a2d0cd5cbd5b96} \\
GUY12 & \url{https://models.physiomeproject.org/exposure/cd10322c000e6ff64441464f8773ed83} \\
GUY13 & \url{https://models.physiomeproject.org/exposure/4c73baa30cb81cf0cd83b4e8d42a5358} \\
GUY14 & \url{https://models.physiomeproject.org/exposure/b897ad1b96031d40293b2e2e10684ffe} \\
HAL01 & \url{https://models.physiomeproject.org/exposure/77bdebf1280135c36f94cac077f74e18} \\
HAT01 & \url{https://models.physiomeproject.org/exposure/c14e77a831ec3f1e56ef03240986a8b7} \\
HAU01 & \url{https://models.physiomeproject.org/exposure/f9b96930ad2e7f5c9e15b1169a9378c7} \\
HEY01 & \url{https://models.physiomeproject.org/exposure/3c875389313cd73f120d69a59cc65acb} \\
HOD01 & \url{https://models.physiomeproject.org/exposure/5d116522c3b43ccaeb87a1ed10139016} \\
HUA01 & \url{https://models.physiomeproject.org/exposure/3227298fc9cc3afc14d058750142c69c} \\
HUN01 & \url{https://models.physiomeproject.org/exposure/f4b7120aa512c7f5e7a0664abcee3e8b} \\
HUN02 & \url{https://models.physiomeproject.org/exposure/f4b7120aa512c7f5e7a0664abcee3e8b} \\
HYN01 & \url{https://models.physiomeproject.org/exposure/b5c9ac7a7a8b76918c02b39967ae191b} \\
HYP01 & \url{https://models.physiomeproject.org/exposure/ea5144be46b69ba503a2d0cd5cbd5b96} \\
INA01 & \url{https://models.physiomeproject.org/exposure/08bcead2dc05cf2709a598e7f61a6182} \\
INA02 & \url{https://models.physiomeproject.org/exposure/08bcead2dc05cf2709a598e7f61a6182} \\
IRI01 & \url{https://models.physiomeproject.org/exposure/7af88a567b3fdbb5209ddd2eee44075c} \\
IZA01 & \url{https://models.physiomeproject.org/exposure/94a5dddfa94d46f1d842e4e87a94646f} \\
JEL01 & \url{https://models.physiomeproject.org/exposure/4817bdfdd245412f52dd79da59137fe8} \\
JEL02 & \url{https://models.physiomeproject.org/exposure/4817bdfdd245412f52dd79da59137fe8} \\
JEL03 & \url{https://models.physiomeproject.org/exposure/4817bdfdd245412f52dd79da59137fe8} \\
KAR01 & \url{https://models.physiomeproject.org/exposure/b5ddf62f58c911e794174d61f9aedf61} \\
KAR02 & \url{https://models.physiomeproject.org/exposure/46a466d519479cee832ebf8b66c1ea00} \\
KAT01 & \url{https://models.physiomeproject.org/exposure/2a3e1e865606c4da94c251017038f820} \\
KAT02 & \url{https://models.physiomeproject.org/exposure/2a3e1e865606c4da94c251017038f820} \\
KOM01 & \url{https://models.physiomeproject.org/exposure/88ec4bc9780f123780b75e4c9ce1682b} \\
KOM02 & \url{https://models.physiomeproject.org/exposure/fa7996b256f19f92568fb7ad414b6cba} \\
KYR01 & \url{https://models.physiomeproject.org/exposure/fd3eeaae0cbb662a9d5a2b47ce81a825} \\
LEL01 & \url{https://models.physiomeproject.org/exposure/1db3a968ce8585201acd753ce857002a} \\
LEL02 & \url{https://models.physiomeproject.org/exposure/1db3a968ce8585201acd753ce857002a} \\
LEM01 & \url{https://models.physiomeproject.org/exposure/3094a5c3e0810028a9accc3b773cab36} \\
LEM02 & \url{https://models.physiomeproject.org/exposure/3094a5c3e0810028a9accc3b773cab36} \\
LEN01 & \url{https://models.physiomeproject.org/exposure/fb612e6ce429bc45d99388c045a100cb} \\
LEN02 & \url{https://models.physiomeproject.org/exposure/520cbff71e195d0a0ed36ce1c78d46d5} \\
LEN03 & \url{https://models.physiomeproject.org/exposure/520cbff71e195d0a0ed36ce1c78d46d5} \\
LI01 & \url{https://models.physiomeproject.org/exposure/dfe4f6c90d58266f0f5d6d320c291e40} \\
LI02 & \url{https://models.physiomeproject.org/exposure/dfe4f6c90d58266f0f5d6d320c291e40} \\
LI03 & \url{https://models.physiomeproject.org/exposure/dfe4f6c90d58266f0f5d6d320c291e40} \\
LIV01 & \url{https://models.physiomeproject.org/exposure/dedfcd1a135ddb59e9d979ec1376a44f} \\
M01 & \url{https://models.physiomeproject.org/exposure/d20c1a1db1b4027605b87b4361e62560} \\
MAC01 & \url{https://models.physiomeproject.org/exposure/a499a7082706164315ad07feff408850} \\
MAC02 & \url{https://models.physiomeproject.org/exposure/1229d793e4a010088736d682bd35d463} \\
MAC03 & \url{https://models.physiomeproject.org/exposure/1229d793e4a010088736d682bd35d463} \\
MAH01 & \url{https://models.physiomeproject.org/exposure/a5586b72d07ce03fc40fc98ee846d7a5} \\
MAI01 & \url{https://models.physiomeproject.org/exposure/a195d957a2d63ac4defe3232fd0ea50c} \\
MAI02 & \url{https://models.physiomeproject.org/exposure/9970597960dbe3a381d773d90c0298c2} \\
MAI03 & \url{https://models.physiomeproject.org/exposure/feccb0f0649425a61526049f35c3b7fb} \\
MAL01 & \url{https://models.physiomeproject.org/exposure/8e081783cbe18d346eaf14cd5c1ae18f} \\
MAL02 & \url{https://models.physiomeproject.org/exposure/3c1c7b17df06921a4e1b05c639a45d32} \\
MCA01 & \url{https://models.physiomeproject.org/exposure/60e23c3228a3e455699846704006a8fe} \\
MCA02 & \url{https://models.physiomeproject.org/exposure/60e23c3228a3e455699846704006a8fe} \\
MIC01 & \url{https://models.physiomeproject.org/exposure/9e77fb28778473706ebd43bd5bea86d7} \\
MIF01 & \url{https://models.physiomeproject.org/exposure/8854e495f7e4240b1607820088b13138} \\
MIJ01 & \url{https://models.physiomeproject.org/exposure/a9cd3fd1ec3c3bbe2613ca9e7490a367} \\
MIT01 & \url{https://models.physiomeproject.org/exposure/0d32bf39c8af51c44373b2e68c3cec74} \\
MIT02 & \url{https://models.physiomeproject.org/exposure/9efc295a6f3360044b4712c28b8314e7} \\
MOD01 & \url{https://models.physiomeproject.org/exposure/c49d416ae3a5132882e6ea7479ba50f5} \\
NAZ01 & \url{https://models.physiomeproject.org/exposure/45de427094d4814e74156890fd99fcc6} \\
NAZ02 & \url{https://models.physiomeproject.org/exposure/45de427094d4814e74156890fd99fcc6} \\
NEL01 & \url{https://models.physiomeproject.org/exposure/e811b8311a550fb0eedf95402b3166d0} \\
NEU01 & \url{https://models.physiomeproject.org/exposure/a7e3d6ea01b67ecbe06e11f7909bcb93} \\
NIE01 & \url{https://models.physiomeproject.org/exposure/59a44249dec83576d97fd3fce46ec5f9} \\
NIE02 & \url{https://models.physiomeproject.org/exposure/97fb1de5199b1a74c89281db97aecc13} \\
NOB01 & \url{https://models.physiomeproject.org/exposure/e42fb39e149f444be872e04e6e731ac0} \\
NOB02 & \url{https://models.physiomeproject.org/exposure/a40c4434423c0436e2789a2d457b7ab2} \\
NOV01 & \url{https://models.physiomeproject.org/exposure/ea5f37b01b27c7ba25b6c179a12a464a} \\
NOV02 & \url{https://models.physiomeproject.org/exposure/1e1bee6ef3243503e7e1531cfd61bb3f} \\
NOV03 & \url{https://models.physiomeproject.org/exposure/e24887f982e9246d05ba0f7152bd4aaa} \\
NYG01 & \url{https://models.physiomeproject.org/exposure/ad761ce160f3b4077bbae7a004c229e3} \\
OST01 & \url{https://models.physiomeproject.org/exposure/d9de93b128da322a4d50f24589980ea1} \\
OVE01 & \url{https://models.physiomeproject.org/exposure/504e7681708fc7b1260db25363658be5} \\
PAR01 & \url{https://models.physiomeproject.org/exposure/a07c660076d3caddac2d136b0c0c7c05} \\
PED01 & \url{https://models.physiomeproject.org/exposure/00bc5ddf4f576af8f81ffc429a183a73} \\
PHI01 & \url{https://models.physiomeproject.org/exposure/87ef0e70902ed8a49d10252617835aa9} \\
PMR01 & \url{https://models.physiomeproject.org/exposure/f2a5a551e8e22f18e5adb641243269c5} \\
PMR01 & \url{https://models.physiomeproject.org/exposure/62b7b4336d5d1caa0a01c3faff8b33ea} \\
PMR01 & \url{https://models.physiomeproject.org/exposure/2fa5de12b0923874aee04486617eb2bd} \\
PMR02 & \url{https://models.physiomeproject.org/exposure/52e2a146e8bf5976379803fec46138a7} \\
PMR02 & \url{https://models.physiomeproject.org/exposure/29a0ec2468a49a64a123f927083260f0} \\
POT01 & \url{https://models.physiomeproject.org/exposure/05510be013e1d096e26c3716c950712d} \\
POT02 & \url{https://models.physiomeproject.org/exposure/80982c99e643a576d10e7f3e271a2299} \\
POT03 & \url{https://models.physiomeproject.org/exposure/80982c99e643a576d10e7f3e271a2299} \\
PRO01 & \url{https://models.physiomeproject.org/exposure/026ccefa76a5ea437d6339aae90c2d37} \\
PUL01 & \url{https://models.physiomeproject.org/exposure/63f124c018745aa8ec92a091434f3dfc} \\
PUR01 & \url{https://models.physiomeproject.org/exposure/baa78a664bd1f620acce7257883fc91a} \\
PUR02 & \url{https://models.physiomeproject.org/exposure/baa78a664bd1f620acce7257883fc91a} \\
RAL01 & \url{https://models.physiomeproject.org/exposure/d8dae90ac67ab8b892864c7c66bf6110} \\
RAN01 & \url{https://models.physiomeproject.org/exposure/9fb32c29b3650b853590d30c7e3374af} \\
RAP01 & \url{https://models.physiomeproject.org/exposure/ea84bd6e27e521677bfc2d6283b371f3} \\
RAT01 & \url{https://models.physiomeproject.org/exposure/f9611ad158302284299c1020faf878ee} \\
RAZ01 & \url{https://models.physiomeproject.org/exposure/6703dcc072ed1c21cade028b9abbcfa0} \\
REE01 & \url{https://models.physiomeproject.org/exposure/b0a3d0dbb029b6cbd2a193014b779a82} \\
REI01 & \url{https://models.physiomeproject.org/exposure/cfbc61b251bcccb085ffd9d074c1d44c} \\
REV01 & \url{https://models.physiomeproject.org/exposure/03852e75c08161cf63d70417ec4a2fae} \\
RIC01 & \url{https://models.physiomeproject.org/exposure/e340f005288ec6bd49535cf792769dd0} \\
ROM01 & \url{https://models.physiomeproject.org/exposure/9a75ea45513cec9ac59991afa40430a4} \\
SCH01 & \url{https://models.physiomeproject.org/exposure/926ad168d96e3fe4b87f452890631f86} \\
SED01 & \url{https://models.physiomeproject.org/exposure/c18a94e82ef7ed7c9dad399e61677b5e} \\
SED02 & \url{https://models.physiomeproject.org/exposure/c18a94e82ef7ed7c9dad399e61677b5e} \\
SHO03 & \url{https://models.physiomeproject.org/exposure/e10def01073e56fb32f06bf59e66a318} \\
SNE01 & \url{https://models.physiomeproject.org/exposure/4f67bf3d82a1ad07d332e383937041e2} \\
SNE02 & \url{https://models.physiomeproject.org/exposure/88053fd5c8867b72ee23c1702097e377} \\
SNY01 & \url{https://models.physiomeproject.org/exposure/8d791fd6aba9793fb0149d5b6e3afcee} \\
SRI01 & \url{https://models.physiomeproject.org/exposure/500b72e21302febd0f3dc746dc22af81} \\
STE01 & \url{https://models.physiomeproject.org/exposure/b060fdbcfae8c7d85e595c24d36ab11b} \\
SVE01 & \url{https://models.physiomeproject.org/exposure/aad2ca9cce58dba12c8d9d85dccada0f} \\
THA01 & \url{https://models.physiomeproject.org/exposure/1985e7c820ff102b1caebc241df65ce7} \\
TRA01 & \url{https://models.physiomeproject.org/exposure/cfa7684fb084ad748bf3061569d99334} \\
UED01 & \url{https://models.physiomeproject.org/exposure/738882ee938d6b2e2d264714259e0416} \\
VAN01 & \url{https://models.physiomeproject.org/exposure/a0df3e434c6d37073783883cb4dff2d3} \\
VAS01 & \url{https://models.physiomeproject.org/exposure/05557c5c5ae347b0a6099f66eb33e50f} \\
VEM01 & \url{https://models.physiomeproject.org/exposure/c1e5f2110188582f6f44f85930733144} \\
VIL01 & \url{https://models.physiomeproject.org/exposure/d5af7e23b0b67c042ca76949c7e59c79} \\
VIL02 & \url{https://models.physiomeproject.org/exposure/882bd564f637b7a506fafdf8a953a506} \\
VIN01 & \url{https://models.physiomeproject.org/exposure/07cdb663179c77de23e425a56464a2aa} \\
VIN02 & \url{https://models.physiomeproject.org/exposure/07cdb663179c77de23e425a56464a2aa} \\
VIS01 & \url{https://models.physiomeproject.org/exposure/4c26896c035fd04e70658d792affad9d} \\
VIS02 & \url{https://models.physiomeproject.org/exposure/4c26896c035fd04e70658d792affad9d} \\
WAN01 & \url{https://models.physiomeproject.org/exposure/a9e94f2a37908bf3ebb1d6aa4bef82c4} \\
WAN02 & \url{https://models.physiomeproject.org/exposure/a93163783c79b766b9ab88c8a20c1e41} \\
WOD01 & \url{https://models.physiomeproject.org/exposure/eb9280eef64afce83878426e24d5b31c} \\
WOD02 & \url{https://models.physiomeproject.org/exposure/9fddc0e3b1bb1d6ad86baeea2e053715} \\
WOD03 & \url{https://models.physiomeproject.org/exposure/869cfba0248523dcd57a7e49185f70ab} \\
WOD04 & \url{https://models.physiomeproject.org/exposure/eb9280eef64afce83878426e24d5b31c} \\
WOL01 & \url{https://models.physiomeproject.org/exposure/c7801370ba2f7c3e19ad4d12d55d4968} \\
WOL02 & \url{https://models.physiomeproject.org/exposure/c1821457cc8c8135f3317ec4882b719f} \\
WOL03 & \url{https://models.physiomeproject.org/exposure/f68c97d153bb0e4a050cb4e143b047a8} \\
YAM01 & \url{https://models.physiomeproject.org/exposure/5ad3bc65fe5ed254806310e2dfb1c5cb} \\
YAT01 & \url{https://models.physiomeproject.org/exposure/0b98c29503e70a6b35be0ed0f5719458} \\
\end{longtable}}

\clearpage\section{Hyperparameter Search}\label{app:hype}
\normalsize
We exclusively conduct hyperparameter search on fold 0. 
For \textbf{GraFITi}~\citep{Yalavarthi2024.GraFITi} the hyperparameters for the search are as follows:
\begin{itemize}
    \item The number of layers, with possible values [1, 2, 3, 4].
    \item The number of attention heads, with possible values [1, 2, 4].
    \item The latent dimension, with possible values [16, 32, 64, 128, 256].
\end{itemize}

For the \textbf{LinODEnet} model~\citep{Scholz2022.Latenta} we search the hyperparameters from:
\begin{itemize}
    \item The hidden dimension, with possible values [16, 32, 64, 128].
    \item The latent dimension, with possible values [64, 128, 192, 256].
\end{itemize}

For \textbf{GRU-ODE-Bayes}~\citep{DeBrouwer2019.GRUODEBayesd} we tune the hidden size from [16, 32, 64, 128, 256]

For \textbf{Neural Flows}~\citep{Bilos2021.Neurald} we define the hyperparameter spaces for the search are as follows:
\begin{itemize}
    \item The number of flow layers, with possible values [1, 2, 4].
    \item The hidden dimension, with possible values [16, 32, 64, 128, 256].
    \item The flow model type, with possible values [GRU, ResNet].
\end{itemize}

For the \textbf{CRU}~\citep{Schirmer2022.Modelingb} the hyperparameter space is as follows:
\begin{itemize}
    \item The latent state dimension, with possible values [10, 20, 30].
    \item The number of basis functions, with possible values [10, 20].
    \item The bandwidth with possible values [3, 10].
\end{itemize}

\clearpage\section{Regularly Sampled Time Series Data Experiments}\label{app:regular}

In this chapter, we show important findings related to generating \textbf{regularly sampled multivariate time series data} with \Bench. 
We discuss the generation process of 3 datasets that were chosen based on 2 important criterions.
First is the previously mentioned JGD which was shown to be important indicator for the complexity of the dataset. 
Second criterion is for these datasets to include a high number of channels as we show through the following small experiment, how channel dependency matters more and have more effect when tested on our ODE-datasets compared to current benchmark datasets.

The datasets used here were DOK01, INA01, and HYN01 which all have a number of channels greater than 15 with 18, 29, and 22 channels respectively. 
Furthermore, they are ranked highly with respect to the JGD metric having values of 2,277, 2,218, and 1,548 respectively which makes them the most complex high dimensional (more than 15 channels) datasets in our benchmark.
Datasets were generated based on the configuration at~\Cref{sec:dsinfo}, generating 100 samples for each dataset based on 100 different initial conditions that are drawn uniformly at random from a distribution based on the configuration given to the model.
Each time series sample spans 300 timesteps. 
The datasets were split into a train/val/test split of 70/20/10 which is similar to the split done normally on the TSF literature for the weather, electricity and traffic datasets~\citep{Chen2023.TSMixerb, Zeng2023.Are, Nie2023.Time}.   

We conduct experiments on state-of-the-art Time Series Forecasting models PatchTST~\citep{Nie2023.Time}, TSMixer~\citep{Chen2023.TSMixerb}, and DLinear~\citep{Zeng2023.Are}. 
For DLinear, there is only a univariate model (doesn't capture channel dependency), but for the other models, they can be implemented as univariate or multivariate (capture channel dependency) variants.
So overall, we label the univariate models as \textbf{CI} (\textbf{C}hannel \textbf{I}ndepedent), and multivariate models as \textbf{CD} (\textbf{C}hannel \textbf{D}ependent). 
These models are tested with the 3 ODE datasets mentioned earlier tuning the hyperparameters for the 3 models through random search of 20 trials for a predefined range of parameters.

For PatchTST, we set the following hyperparameter range:

\begin{itemize}
	
	\item learning rate (range of floats) $\rightarrow$ [$10^{-7}$, $10^{-2}$].
	\item number of encode layers (range of integers) $\rightarrow$ [1, 6].
	\item Dimensionality of fully connected layer (categorical) $\rightarrow$ [256, 512, 1024].
	\item Embedding dimension of the attention layer (categorical) $\rightarrow$ [128, 256, 512, 1024].
	\item dropout (range of floats) $\rightarrow$ [0, 0.9].
	\item fully connected layer dropout (range of floats) $\rightarrow$ [0, 0.9].
	\item patch size (categorical) $\rightarrow$ [4, 8].
	\item stride (categorical) $\rightarrow$ [2, 4].
\end{itemize}

For TSMixer, we set the following hyperparameter range:

\begin{itemize}
	
	\item learning rate (range of floats) $\rightarrow$ [$10^{-7}$, $10^{-2}$].
	\item number of blocks (range of integers) $\rightarrow$ [1, 5].
	\item hidden size of MLP layers (categorical) $\rightarrow$ [32, 64, 256, 1024].
	\item dropout (range of floats) $\rightarrow$ [0, 0.9].
	\item activation (categorical) $\rightarrow$ [ReLU, GeLU].
	
\end{itemize}

For DLinear, learning rate was tuned on the same range of values mentioned above.
All models were tuned on the validation set. 
The results for the best performing model on validation set (lowest MSE error) were reported on the test set.
We used a lookback window of 24 and forecasting horizon of 12 for all the conducted experiments on the ODE datasets.
On the other hand, we report the results of the mentioned models on  well-established evaluation datasets such as ETTh1, ETTh2, ETTm1, ETTm2, Weather, Electricity, and Traffic on a forecasting horizon of 96. 
The reported results from~\Cref{tab:regular_time_series_forecasting_results} give us interesting insights about our benchmark on the regularly sampled time series data.

\begin{table}[!h]
	\centering
	\scriptsize
	\begin{threeparttable}
		
		\caption{%
			MSE Test loss reported on 3 ODE datasets as well as on 7 of the most popular regular TSF datasets.
			We highlight the best performing version (CI/CD) of each model in \textbf{bold}.
			The results were reported on 5 different baselines showing a clear trend of CD models being better on our generated ODE datasets.
			On the other hand, for benchmark datasets, this trend is not there as CI models are performing quite well.
		}\label{tab:regular_time_series_forecasting_results}
		
		\begin{tabular}{l ccccc}
			\toprule
			Dataset & PatchTST(CI) & PatchTST(CD) & TSMixer(CI) & TSMixer(CD) & DLinear
			\\  \midrule
			INA01 & 0.099 & \textbf{0.052} & 0.168 & \textbf{0.062} & 0.191 \\ 
			DOK01 & 0.087 & \textbf{0.022} & 0.155 & \textbf{0.039} & 0178 \\
			HYN01 & \textbf{0.001} & \textbf{0.001} & 0.019 & \textbf{0.002} & 0.046 \\
			\hline
			ETTh1 & \textbf{0.375*} & 0.416* & 0.361* & \textbf{0.359*} & 0.375* \\
			ETTh2 & \textbf{0.274*} & 0.334* & \textbf{0.274*} & 0.275* & 0.289* \\
			ETTm1 & \textbf{0.290*} & 0.326* & 0.285* & \textbf{0.284*} & 0.299* \\
			ETTm2 & \textbf{0.165*} & 0.195* & 0.163* & \textbf{0.162*} & 0.167* \\
			Weather & \textbf{0.152*} & 0.168* & \textbf{0.145*} & \textbf{0.145*} & 0.176* \\
			Electricity & \textbf{0.130*} & 0.196* & \textbf{0.131*} & 0.132* & 0.140* \\
			Traffic & \textbf{0.367*} & 0.595* & 0.376* & \textbf{0.370*} & 0.410* 
			\\  \bottomrule
		\end{tabular}
		\begin{tablenotes}
			\small
			\item[*] Results are taken directly for PatchTST, TSMixer, and DLinear from the respective papers \citep{Nie2023.Time}, \citep{Chen2023.TSMixerb}, and \citep{Zeng2023.Are}. 
		\end{tablenotes}
		
	\end{threeparttable}
\end{table}

We notice from the results how unpredictable the results can be for the benchmark datasets, with \textbf{CI} models being better on some datasets and \textbf{CD} models being better on the rest. 
Also, one more interesting finding is how close the test loss is between a very simple CI model such as DLinear and the state-of-the-art models which can be concerning regarding how much channel dependency there is on the available benchmark datasets. 
On the other hand, for our ODE-generated datasets, the results are more consistent with clear better performance for \textbf{CD} models over their \textbf{CI} counterparts.
Also, the best model overall was noticed to be always a \textbf{CD} model which shows the importance of capturing channel dependency on the ODE generated datasets.
One more interesting finding is how worse is a DLinear model on our benchmark datasets where its results are at least 5 times worse than the best performing model.
That shows how complex these datasets are, as well as how multivariate models can clearly benefit from intrinsic relation between the channels on such datasets. 
\clearpage\section{Lorenz}\label{app:lorenz}
We conducted a forecasting experiment on the chaotic Lorenz Attractor: 
\begin{align}
    \frac{dx}{dt} &= \sigma (y - x) \\
    \frac{dy}{dt} &= x (\rho - z) - y \\
    \frac{dz}{dt} &= xy - \beta z
\end{align}

Instead of varying the constants as we do in \Bench, we set the parameters to 
\begin{itemize}
    \item $\rho = 28$
    \item $\sigma = 10$
    \item $\beta=\frac{8}{3}$
\end{itemize}
We vary the sample the initial states from $x \sim [1,3], y\sim[0,2] z\sim[0,2]$.
Following \citet{Gilpin2021.Chaosa}, the task is to forecast the final $\frac{1}{6}$ of the IMTS based on the initial $\frac{5}{6}$.
Since we always use the exact same constants, we create 200 instead of 2000 time series instances. 
Outside the mentioned changes, the experimental protocol is the same as the one used in our experiments with the ODEs from \Bench.

As shown in \Cref{tab:lorenz}, there is no model which significantly outperforms the constant baseline  GraFITi-C.

\begin{table}[!h]
    \scriptsize
    \caption{%
        Test MSE IMTS created on the chaotic Lorenz-Attractor.
    }\label{tab:lorenz}
    \centering
    \begin{tabular}{l cccccc}
        \toprule
    \\	Dataset & GRU-ODE & LinODEnet & CRU & Neural Flow & GraFITi & GraFITi-C
    \\	\midrule
        Lorenz & 1.334±0.097 & 1.313±0.123 & 1.292±0.087 & 1.345±0.105 & 1.294±0.090 & 1.303±0.079
    \\	\bottomrule
    \end{tabular}
\end{table}

\end{document}